%% file: main.tex
\pdfoutput=1
\documentclass{article}
\usepackage{preprint,times}

\input{math_commands.tex}

\PassOptionsToPackage{table}{xcolor}
\usepackage{hyperref}
\usepackage{url}
\usepackage{booktabs}
\usepackage{listings}
\usepackage[table]{xcolor}
\usepackage{multirow}
\usepackage{amsmath}
\usepackage{adjustbox}
\usepackage{makecell}
\usepackage{enumitem}
\usepackage{graphicx}
\hypersetup{colorlinks=true, linkcolor=red!60!black, citecolor=blue!60!black, urlcolor=blue!60!black}

\lstdefinestyle{prompt}{
    basicstyle=\ttfamily\scriptsize,
    breaklines=true,
    breakatwhitespace=false,
    frame=single,
    backgroundcolor=\color{black!5},
    xleftmargin=2pt,
    xrightmargin=2pt,
    aboveskip=4pt,
    belowskip=4pt,
}

\title{Back to the Question: Tracing Visual\\ Distraction Through Text Tokens}

\author{
Marie Amale Huynh\textsuperscript{1} \quad
Regev Cohen\textsuperscript{2} \quad
Serena Yeung-Levy\textsuperscript{1}\thanks{Equal advising.} \quad
Sivan Doveh\textsuperscript{1}\footnotemark[1]
\\[0.5em]
\textsuperscript{1}Stanford University \quad
\textsuperscript{2}Reichman University
\\[0.3em]
\texttt{mahuynh@stanford.edu}
}

\preprintfinalcopy

\renewcommand{\headrulewidth}{0pt}
\renewcommand{\footrulewidth}{0pt}

\begin{document}

\maketitle

\begin{abstract}

The visual world is cluttered: robust vision-language reasoning requires ignoring irrelevant information. Yet vision-language models (VLMs) can suffer substantial accuracy losses when answering questions about a target image with distractors present. Existing mitigations focus on selecting relevant images or limiting information exchange between images. Through a systematic analysis of three open 7–8B VLMs on five perception benchmarks adapted to include distractor images, we show that distractors reduce accuracy even when models correctly identify the relevant image, suggesting that selection alone is not enough. To understand the mechanism behind this interference, we combine attention analysis with targeted knockouts in Qwen2.5-VL and identify middle-layer distractor-to-text connections as a pathway through which irrelevant visual information affects reasoning. We introduce answer consistency regularization, which encourages consistent answer distributions with and without distractors. Without directly constraining attention, this objective reduces middle-layer attention from question tokens to distractor images, consistent with the pathway our analysis implicates. With five distractors, our approach improves circular accuracy by an average of 16 percentage points over the Qwen2.5-VL baseline across the five benchmarks while preserving single-image performance.

\end{abstract}

\section{Introduction}

Distraction tests true understanding: exams hide key variables among irrelevant data; interviews frame critical questions within long stories. The ability to reason despite distractions is essential for deploying vision-language models (VLMs) in real-world environments, where visual inputs are cluttered and noisy. Although VLMs excel on isolated single-image benchmarks, performance degrades sharply when irrelevant images are introduced \citep{sharma2024losing, das2026more, wu2025visual, liu2025robustness}, exposing a crucial gap between evaluation and deployment.

\begin{figure}[h]
    \centering
    \includegraphics[width=\linewidth]{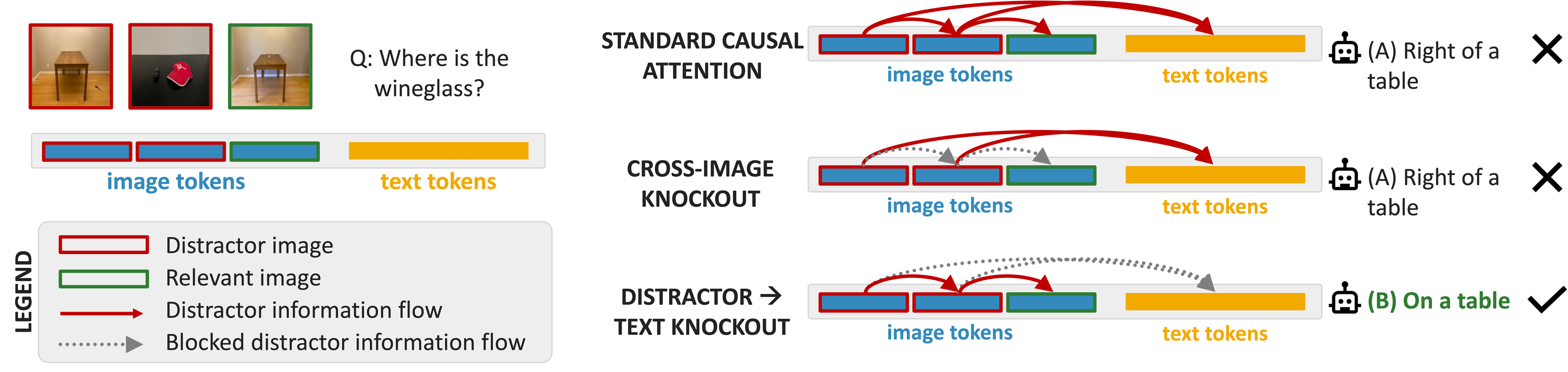}
    \vspace{-0.6cm}
    \caption{\textbf{How does distractor information reach the answer?} Cross-image knockout leaves distractor-to-text connections open. In this example (\emph{Where is the wineglass?}), standard attention (top) and cross-image knockout (middle) yield incorrect answers, while distractor-to-text knockout (bottom) restores the correct answer. Knockouts target middle layers. Outlines identify distractor (red) and relevant (green) images. Arrows show allowed (solid red) and blocked (dotted gray) information flow from distractors; other connections are omitted.}

    \label{fig:attention_knockout}
\end{figure}

Existing mitigations filter irrelevant images before interference \citep{wu2025visual}, restrict information exchange between images \citep{das2026more}, or discourage answers grounded in the wrong image \citep{liu2025mia}. However, these interventions reduce but do not eliminate distractor interference. A possible source of residual interference is a text-mediated pathway through the model's internal question representations. In single-image description settings, \citet{kaduri2025s} show that input query-text representations encode global image context and can support response generation even when generated tokens cannot directly attend to image tokens. Masks that restrict attention between images \citep{das2026more} leave this potential pathway open, since question tokens can still attend to distractor images (Figure~\ref{fig:attention_knockout}). Whether distractors exploit this text-mediated pathway to interfere with reasoning remains an open question.

Answering a question in the presence of distractor images requires both identifying the relevant image and reasoning about it. To examine how distractors affect these two demands, we adapt five perception benchmarks (CLEVR \citep{johnson2017clevr}, CV-Bench \citep{tong2024cambrian}, MIMIC \citep{das2026more}, What'sUp \citep{kamath2023s}, and BLINK \citep{fu2024blink}) to include distractor images, and evaluate three open 7–8B VLMs (Qwen2.5-VL-7B \citep{bai2025qwen25vltechnicalreport}, InternVL3-8B \citep{zhu2025internvl3}, and LLaVA-OneVision-7B \citep{li2024llava}). We compare three inference regimes: standard prompting; two-stage prompting, in which the model first selects an image and then answers with all images retained; and prompting that explicitly specifies the target image. These comparisons probe two failure modes: \emph{selection failure}, where the model selects the wrong image, and \emph{in-context corruption}, where distractors reduce accuracy even when the target is explicitly specified. Notably, with five distractors, explicitly naming the target image recovers none of the drop in circular accuracy across the three models and five benchmarks, showing that distractor interference persists well beyond selection errors.

Our knockout experiments (Figure~\ref{fig:attention_knockout}) show that blocking distractor information from entering text representations repairs significantly more distraction-induced failures than restricting inter-image attention. In Qwen2.5-VL with five distractors, severing distractor-to-text connections in layers 9–17 recovers 47\% of failures while disrupting only 5\% of correctly answered examples. In contrast, inter-image masking in the same layers recovers just 10\% of failures while disrupting 57\% of correct outputs. Attention analysis confirms that question tokens attend heavily to distractor images in middle layers—a pathway left entirely unblocked by inter-image masking. Together, these findings support text representations as an important pathway for distractor interference that inter-image masking leaves open.

To reduce distractors' effect on model predictions, we introduce answer consistency regularization. During fine-tuning, we minimize the Kullback–Leibler (KL) divergence between the model’s answer distributions for the same question with and without distractor images, using its single-image prediction as the reference. This objective encourages consistent answers across the two contexts without explicitly constraining attention. Our attention analysis further shows that fine-tuning reduces question-token attention to distractor images in the middle layers, where knockouts identified a pathway for distractor interference.

Our contributions are threefold:
\begin{enumerate}[nosep,leftmargin=*]
    \item We distinguish selection failure from in-context corruption through comparisons that vary target guidance and which images remain in context, showing that selecting the right image does not eliminate distractor interference.
    \item We identify middle-layer distractor-to-text attention as a key pathway for in-context corruption. Blocking it recovers 47\% of distraction-induced failures, versus 10\% for inter-image masking.
\item To reduce in-context corruption, we introduce answer consistency regularization, which aligns answers under distraction with single-image predictions. On Qwen2.5-VL with five distractors, it improves circular accuracy by 16 points, with similar gains when the target is named, and reduces middle-layer attention to distractors along the identified pathway.
\end{enumerate}

\section{Related Work}

\paragraph{Multi-Image and Multimodal Reasoning.} 
Multi-image benchmarks focus on evaluating whether VLMs can integrate information across separate visual inputs \citep{suhr2019nlvr2, ying2024mmt, meng2025mmiu}. Despite progress, frontier models frequently rely on superficial shortcuts rather than actual multimodal integration \citep{zhao2024benchmarking, fu2024blink, wang2025muirbench}. Success on these benchmarks does not necessarily establish robustness to irrelevant visual context. Related work documents \textit{cross-image information leakage} as a primary failure mode, where visual representations across images become entangled during processing \citep{park2025mitigating}. Building on these observations, we study robustness to irrelevant visual context by augmenting perception benchmarks with distractor images, retaining a single query-relevant target.

\paragraph{Robustness to Visual Distractions.}
Prior studies show that VLMs lose accuracy as the number of distractors increases, while remaining sensitive to context length and image positioning \citep{sharma2024losing, wu2025visual}. Using MIMIC, \citet{das2026more} examine how relevant information is distributed across images and how visual sequence length affects performance, showing that reducing the number of visual tokens can improve counting accuracy. Our analysis extends this investigation across benchmarks, revealing that sensitivity to distractor content varies widely: while MIMIC is largely resilient, other benchmarks show far greater performance drops. To examine what underlies these performance losses, we also compare inference regimes that vary whether the target image is explicitly identified and whether distractors are included in context. This enables us to distinguish \emph{selection failure} from \emph{in-context corruption}.

\paragraph{Mitigation Methods.} 
Prior approaches address visual distraction through \emph{context reduction} or by improving how models process multiple images in context. Context-reduction methods use query-aware retrieval and token compression to limit the visual context supplied to the language model \citep{wu2025visual}. Other approaches intervene within the model. Delimiter Token Scaling (DTS) strengthens the hidden states of image-delimiter tokens at selected layers, reinforcing image-wise attention boundaries and reducing cross-image information leakage without additional training \citep{lee2026enhancing}. Inter-image attention masking instead restricts information exchange between image tokens during LoRA fine-tuning \citep{das2026more}. Training-based approaches also include MIA-DPO, which constructs distractor-augmented preference pairs for direct preference optimization \citep{liu2025mia}. Our complementary approach aligns prediction distributions between distracted and clean inputs. This enforces distractor robustness without restricting internal information flow, preserving single-image performance.

\paragraph{Mechanistic Analysis of Visual Information Flow.}
Mechanistic analyses of VLMs examine how visual information enters and influences textual representations. In single-image settings, \citet{kaduri2025s} use attention knockouts to show that query-token representations store global image information, while fine-grained visual details are accessed directly from image tokens. They further show that middle layers are central to this cross-modal information transfer. These findings motivate a complementary question: what internal process fails when distractors are present? Through attention analysis and targeted knockouts in Qwen2.5-VL, we find that blocking middle-layer distractor-to-text connections reduces distraction-induced errors, supporting their contribution to in-context corruption.

\section{Experimental Setup}
\label{sec:experimental_setup}

Our experiments examine why irrelevant images impair visual reasoning by considering both the model’s ability to identify and use relevant evidence and the internal pathways through which distractors interfere. We construct controlled multi-image inputs from questions answerable with a single image, keeping the question and required evidence fixed. This setup supports complementary behavioral and mechanistic analyses. Inference comparisons test whether target guidance and distractor removal resolve errors. Attention analysis and targeted knockouts examine which internal connections contribute to interference.

\paragraph{Task Formulation.}\label{sec:task_formulation}
Each example contains a multiple-choice question $q$, a ground-truth answer $a$, and an image set $\mathcal{I}$. Exactly one image $I^* \in \mathcal{I}$ provides the required evidence; the others are distractors. We compare this multi-image input with its single-image counterpart containing only $I^*$, isolating the effect of irrelevant visual context.

\paragraph{Datasets.}
\label{sec:datasets}
We evaluate 9,052 questions from five VQA datasets spanning spatial reasoning, counting, and perception (Table~\ref{tab:dataset_composition}). We adapt each dataset to contain one relevant image and $k \in \{0, 1, 2, 3, 5\}$ distractors. The multiple-choice question, relevant image, and ground-truth answer remain fixed across conditions, allowing us to measure how performance changes as irrelevant visual context increases.

\begin{table}[h]
\small
  \centering
  \caption{\textbf{Composition of the distraction test sets.} Each benchmark is evaluated at 0, 1, 2, 3 and 5 distractors; $n$ is the number of questions, identical at every distraction level.}
  \label{tab:dataset_composition}
  \begin{tabular}{lllr}
    \toprule
    \textbf{Benchmark} & \textbf{Competency} & \textbf{Subsets} & \textbf{$n$} \\
    \midrule
    CLEVR \citep{johnson2017clevr}     & Size reasoning  & size & 3,554 \\
    CV-Bench \citep{tong2024cambrian}  & Vision-centric spatial     & count, relation, depth, distance & 2,560 \\
    What'sUp \citep{kamath2023s}       & Spatial relations          & controlled\_a, controlled\_b & 820 \\
    BLINK \citep{fu2024blink}          & Low-level perception       & \makecell[l]{spatial relation, relative reflectance,\\ relative depth, localization, counting} & 618 \\
    MIMIC \citep{das2026more}          & Counting  & count & 1,500 \\
    \midrule
    \textbf{Total} & & & \textbf{9,052} \\
    \bottomrule
  \end{tabular}
\end{table}

\paragraph{Multi-Image Distractor Construction.}
\label{sec:multi_image_distractor}
To test robustness to visually plausible but irrelevant context, we sample distractors from the same dataset when available (CLEVR, What'sUp, CV-Bench) and use domain-matched images for BLINK. MIMIC already pairs each question with its own distractor images, so we reuse these native distractors rather than sampling new ones. For the other benchmarks, we screen candidates using ground-truth annotations when available or VLM-based answerability checks, rejecting images judged capable of independently supporting an answer to the question. This screening aims to preserve the original answer and prevent distractors from introducing conflicting evidence. We combine accepted distractors with the target image while keeping the question and answer choices unchanged, allowing us to measure the effect of additional visual context without changing the required evidence. Figure~\ref{fig:benchmark_augmentation} illustrates the construction process; dataset-specific details are provided in Appendix~\ref{app:augmentation}.

\begin{figure}[h]
    \centering
    \includegraphics[width=0.9\linewidth]{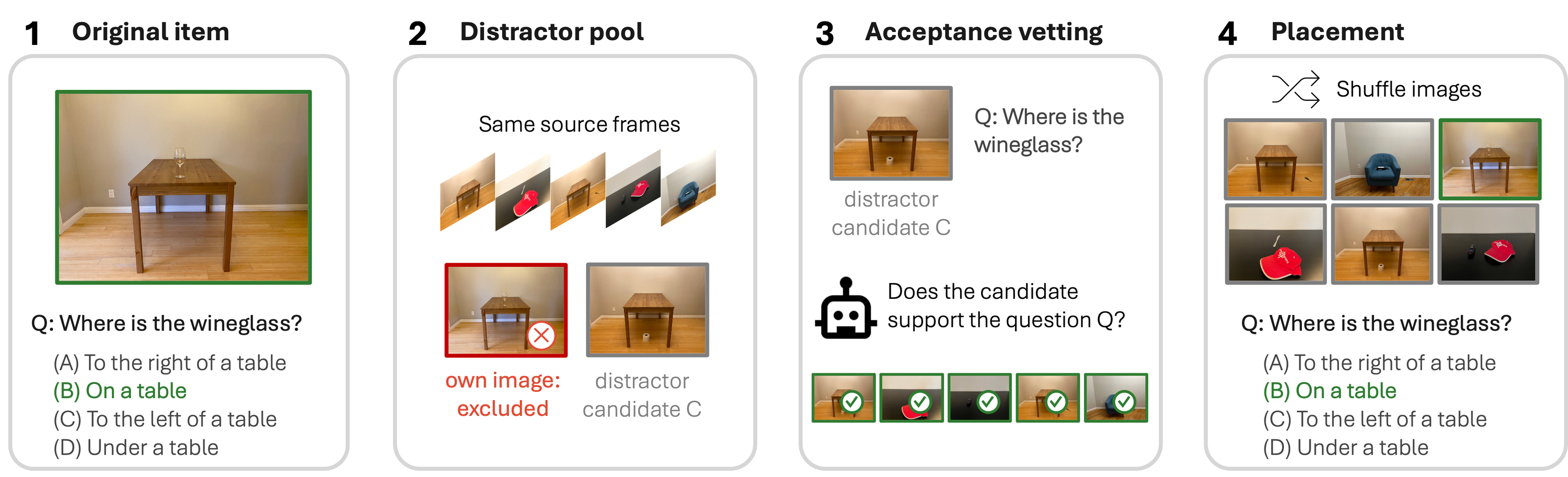}
    \vspace{-0.4cm}
    \caption{\textbf{Distractor augmentation}, illustrated on a What'sUp example. (1) A target image (green) provides the evidence for the question. (2) Candidate distractors are sampled from the same source, excluding the target image. (3) Candidates judged capable of independently supporting an answer are discarded. (4) Accepted distractors are shuffled with the target, keeping the question and answer choices unchanged.}
    \label{fig:benchmark_augmentation}
\end{figure}

\paragraph{Metrics.}
To assess whether answers remain correct across option positions, we follow CircularEval \citep{liu2024mmbench}. For example $i$ with $K_i$ answer choices, we evaluate all $K_i$ circular rotations of the options, placing the correct answer in each position once. Let $c_{i,r}\in \{0,1\}$ indicate whether the model answers example $i$ correctly under rotation $r$. For an evaluation set of $N$ examples, we report mean accuracy across rotations and circular accuracy:

\vspace{-0.5cm}

\begin{equation}
\mathrm{Acc}_{\text{mean}} = \frac{1}{N}\sum_{i=1}^{N} \frac{1}{K_i} \sum_{r=1}^{K_i} c_{i,r},
\qquad
\mathrm{Acc}_{\text{circ}}
= \frac{1}{N}\sum_{i=1}^{N}
\prod_{r=1}^{K_i} c_{i,r}.
\end{equation}

\section{Visual Distractor Interference Persists After Correct Selection}
\label{sec:evaluation}

Does visual distraction primarily prevent models from identifying the relevant image, or does irrelevant context also impair answering after successful selection? We address this question by quantifying the degradation, separating target selection from answering, and using black-image controls to test whether sequence length alone explains the effect.

\begin{figure*}[h]
    \centering
    \includegraphics[width=\linewidth]{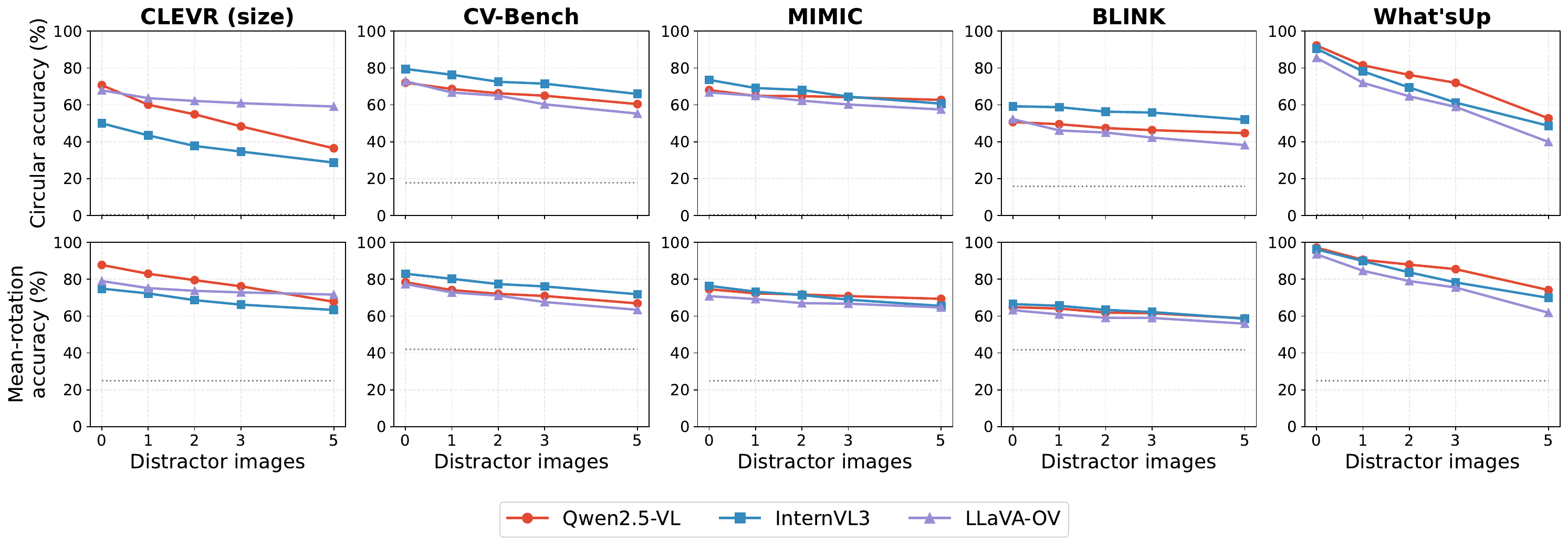}
    \vspace{-0.8cm}
    \caption{\textbf{Accuracy vs. number of distractor images.} All three 7-8B VLMs show degradation, with benchmark-dependent severity.}
    \label{fig:degradation}
\end{figure*}

\paragraph{Mean-rotation accuracy drops by 4 points with one distractor and 13 points with five.} 
Across all three models, performance declines as the number of distractors $k$ increases (Figure~\ref{fig:degradation}). Macro-averaged across models and benchmarks, mean-rotation accuracy drops by $3.7$ percentage points with one distractor and $13.3$ points with five, although the magnitude varies considerably across benchmarks. The stricter circular metric reveals even larger losses in some settings. Qwen2.5-VL's circular accuracy on CLEVR falls from $70.7\%$ without distractors to $36.4\%$ with five.

We compare three inference regimes to test whether identifying the relevant image is sufficient to overcome distraction:

\begin{itemize}[nosep,leftmargin=*]
    \item \textit{Regular:} The model answers directly from all images without target guidance.
    \item \textit{Ground-Truth:} The model answers from all images after the prompt identifies the relevant image.
    \item \textit{Two-Step:} The model predicts the relevant image in phase 1, then answers from all images in phase 2 with its predicted selection identified in the prompt.
\end{itemize}

\paragraph{Selection failures depend strongly on the model.}
Selection accuracy decreases as distractors accumulate but remains above the $1/(k+1)$ chance rate in all evaluated conditions (Table~\ref{tab:image_selection_v3_twostep_rnd}). On CLEVR, accuracy from one to five distractors remains nearly perfect for Qwen2.5-VL ($99.9\%\rightarrow99.4\%$), but falls from $99.9\%$ to $84.1\%$ for InternVL3 and from $93.4\%$ to $29.8\%$ for LLaVA-OV. At five distractors, Qwen2.5-VL still exceeds $81\%$ on every benchmark, whereas LLaVA-OV reaches only $26$-$34\%$. Thus, selection failure is a major source of degradation for some models, but cannot explain Qwen2.5-VL's substantial performance loss.

\input{tables/tab_image_selection_twostep_rnd}

\begin{figure*}[h]
    \centering
    \includegraphics[width=\linewidth]{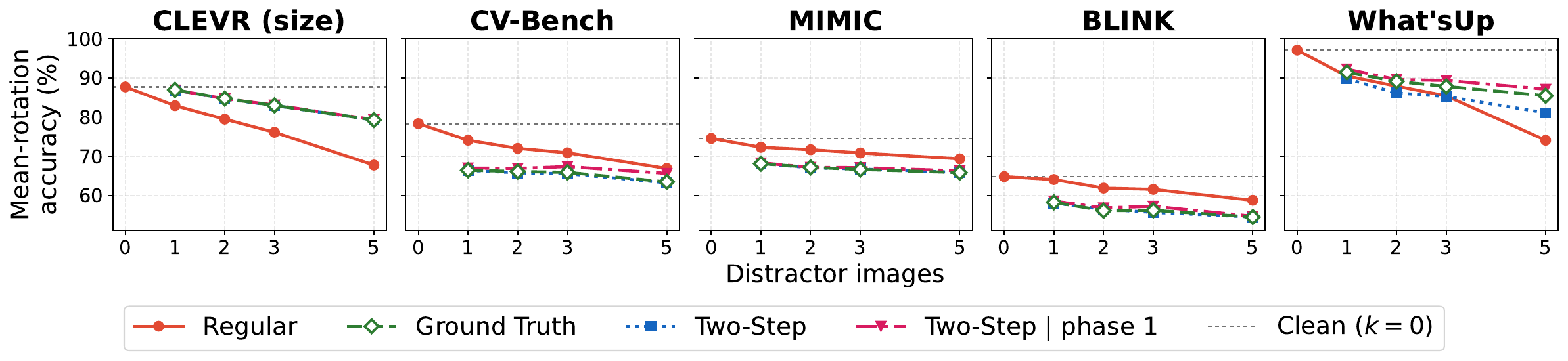}
    \vspace{-0.8cm}
    \caption{\textbf{Target guidance does not eliminate distraction in Qwen2.5-VL.} Mean-rotation accuracy across distractor counts under \textit{Regular}, \textit{Ground-Truth}, and \textit{Two-Step} prompting. We report \textit{Two-Step} both end-to-end, including phase-1 selection errors, and conditioned on correct phase-1 selection. Dashed lines indicate clean accuracy.}
    \label{fig:strategies}
\end{figure*}

\paragraph{Correct target selection recovers at most a quarter of the loss.} Whether the target image is named in the prompt or selected by the model itself, the performance deficit relative to clean single-image inputs largely persists (Figure~\ref{fig:strategies}). For Qwen2.5-VL with five distractors, against a $13.1$-point mean-rotation drop under \textit{Regular} prompting, \textit{Ground-Truth} target naming recovers only $2.3$ points ($18\%$), while \textit{Two-Step} selection recovers just $1.4$ points ($11\%$). Even conditioning \textit{Two-Step} evaluation on correct phase-1 selection recovers only 25\% of the degradation (Figure~\ref{fig:strategies}). Thus, neither specifying the target location nor verifying correct selection restores clean performance. By contrast, removing unselected images after correct selection fully restores accuracy, showing that distractors continue to interfere as long as they remain in context. Full per-benchmark results for Qwen2.5-VL under all prompts and metrics are in Appendix~\ref{app:prompting_strategies_qwen}. The same analysis for InternVL3 and LLaVA-OV appears in Appendix~\ref{app:prompting_strategies}.

\vspace{-0.3cm}

\paragraph{Sequence length.} Distractors add both visual content and sequence length. To separate these effects, we replace real distractors with solid black images of identical dimensions, preserving the number and positions of visual tokens. Under this matched-length control, mean-rotation accuracy drops by only $0.9$ points with one distractor and $3.2$ points with five, compared with $3.7$ and $13.3$ points for real distractors (Figure~\ref{fig:degradation_black}). Thus, at five distractors, competing visual content accounts for an additional $10.1$-point loss---approximately $76\%$ of the total degradation. The contribution of sequence length varies across benchmarks: it accounts for a larger share of the distraction penalty on MIMIC, while content drives most of the degradation on the other benchmarks (Table~\ref{tab:content_fraction_v3}).

Together, these results separate selection failure from in-context corruption: distractors impair answering even when the relevant image is correctly identified. Improving selection alone therefore cannot close the gap.

\section{Distractor Content Contaminates Text Representations}

If selection is not the bottleneck, where does the interference arise? We analyze attention in Qwen2.5-VL-7B and use targeted knockouts to determine which connections contribute to distraction-induced errors. We focus on text-token representations as a potential pathway.

\paragraph{Distractors redirect question-token attention away from the relevant image.} 

\begin{figure}[h]
    \centering
    \includegraphics[width=\linewidth]{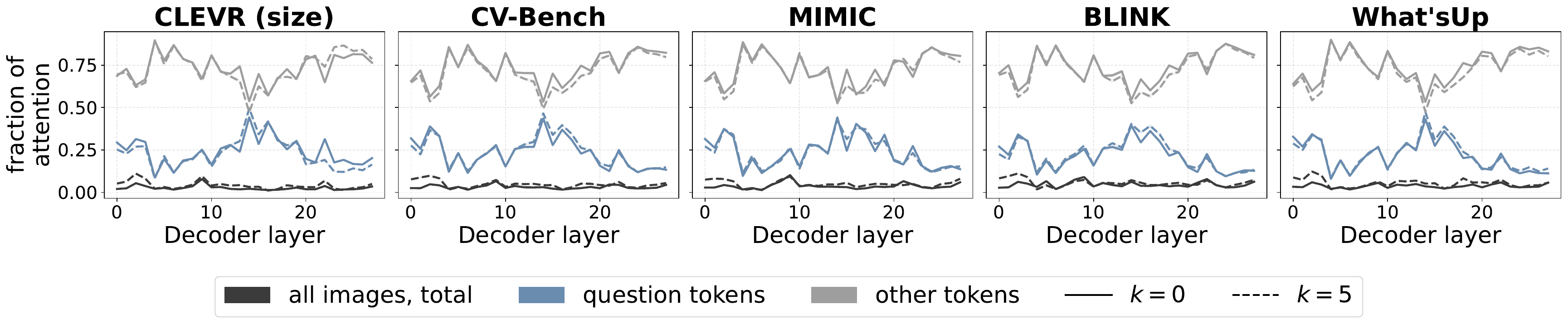}
    \vspace{-0.7cm}
\caption{\textbf{Readout-token attention across decoder layers in Qwen2.5-VL-7B-Instruct.} Curves show the fraction of attention allocated to image tokens (black), question tokens (blue), and remaining tokens (gray; including template tokens, instructions, and image separators). Solid lines denote clean inputs ($k{=}0$), and dashed lines denote inputs with five distractors ($k{=}5$). Results use $n{=}150$ examples per benchmark, except What'sUp ($n{=}300$).}
    \label{fig:readout_targets}
    \vspace{-0.2cm}
\end{figure}

We examine how the answer readout and question tokens distribute attention across textual and visual inputs. Token-group details are provided in Appendix~\ref{app:attention_analysis}. Across all benchmarks the readout attends predominantly to text, placing more than $90\%$ of its attention there and never more than $13\%$ on image tokens at any layer, with its attention to the question peaking sharply at layer $14$ ($46\%$; Figure~\ref{fig:readout_targets}). Introducing five distractor images barely alters this profile: despite a sixfold increase in image tokens, total image attention rises only from $3.6\%$ to $5.1\%$ averaged over layers, and the increase is confined to the first four layers ($3.8\%$ to $8.7\%$).

\begin{figure}[h]
    \centering
    \includegraphics[width=\linewidth]{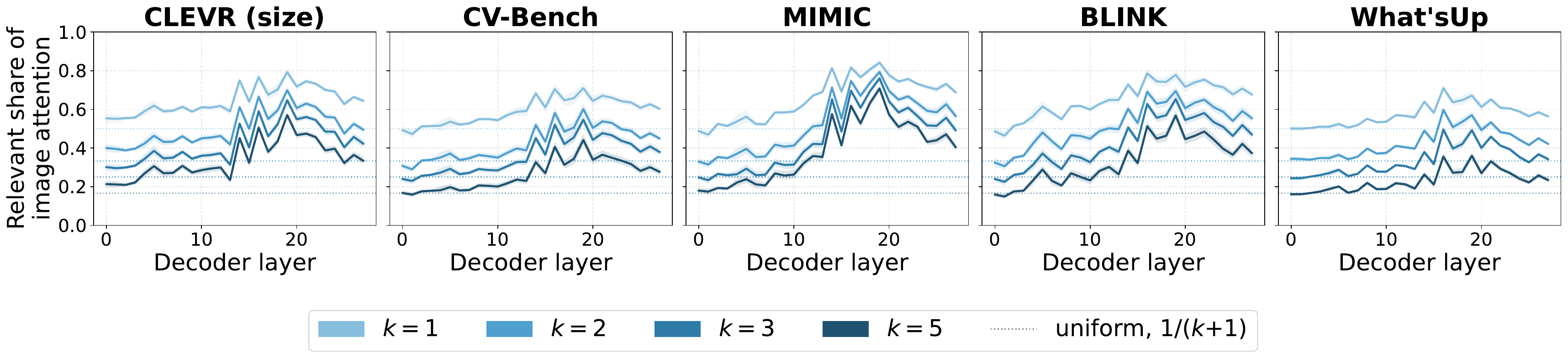}
    \vspace{-0.7cm}
\caption{\textbf{Share of the question tokens' image attention on the relevant image}, $\mathrm{rel}/(\mathrm{rel}+\sum\mathrm{distractors})$, by decoder layer in Qwen2.5-VL-7B-Instruct, at $k = 1, 2, 3, 5$ distractors; dotted lines are uniform allocation $1/(k{+}1)$.  ($n=150$ per benchmark; 300 for What'sUp)}
\label{fig:question_share}
\vspace{-0.5cm}
\end{figure}

Question tokens, in turn, allocate about 10-15\% of their total attention to images. Although this proportion changes little with distractor count, its distribution across images shifts substantially (Figure~\ref{fig:question_share}). Averaged across layers, the target's share of image attention falls from $0.57$-$0.66$ with one distractor to $0.23$-$0.39$ with five. The target remains above the uniform reference $1/(k+1)$ beyond the earlier layers, but distractors collectively receive most of the question tokens' image attention at $k=5$. These observations motivate testing whether the same image-to-text pathway contributes to interference when irrelevant images are present.

\paragraph*{Attention knockouts.}
\label{sec:knockouts}
We test the distractor-to-text pathway by setting selected entries of the additive attention mask to $-\infty$ during prefill, thereby blocking specific connections within designated layer bands. Arrows below denote information flow from source key/value tokens to receiving query tokens. We compare four interventions:
\vspace{-0.1cm}
\begin{description}[nosep, leftmargin=2.2em, font=\normalfont\bfseries]
    \item[P] distractor $\rightarrow$ everything (full-deletion control).
    \item[D] distractor $\rightarrow$ text (primary hypothesis).
    \item[MM] all preceding $\rightarrow$ image (MIMIC cross-image mask \citep{das2026more}).
    \item[R] distractor $\rightarrow$ readout token.
\end{description}

We evaluate each intervention on two subsets. \emph{Broken} examples are answered correctly without distractors but incorrectly under circular evaluation with five distractors; recovery is the fraction made correct by the intervention. \emph{Preserved} examples are correct both with and without distractors under circular accuracy; damage is the fraction made incorrect by the intervention.

In layers $9$--$17$, blocking distractor-to-text attention (\textbf{D}) recovers $47\%$ of broken examples (95\% CI $[45,49]$), approaching the $56\%$ recovered by blocking all distractor connections (\textbf{P}; $[54,58]$). The two interventions largely repair the same failures: $92\%$ of the examples recovered by \textbf{D} are also recovered by \textbf{P} (Table~\ref{tab:knockout-overlap}). Both damage only $5\%$ of preserved examples (Table~\ref{tab:knockout-pooled}). By contrast, the MIMIC-style cross-image mask (\textbf{MM}) \citep{das2026more} recovers only $10\%$ of broken examples ($[9,12]$) while damaging $57\%$ of preserved examples ($[55,58]$). Blocking direct distractor-to-readout attention (\textbf{R}) is less destructive but recovers only $3\%$ ($[3,4]$). Together, these results identify distractor-to-text attention as an effective and selective pathway for repairing in-context corruption (Table~\ref{tab:knockout-pooled}).


\input{tables/tab_knockout_ci_middle_pooled}

\vspace{-0.2cm}

\paragraph{Layer dependence.}
The effect of blocking distractor-to-text attention depends on where the intervention is applied. Masking layers $9$-$17$ recovers 47\% of broken examples, compared with 57\% ($[55,59]$) when masking all 28 layers. These nine middle layers therefore achieve about $82\%$ of the recovery obtained by the full-depth intervention. Per-benchmark recovery ranges from 8-33\% for early-layer masking ($0$-$8$) and 5-13\% for late-layer masking ($18$-$27$), with detailed results in Appendix~\ref{app:knockout}. The strong effect of middle-layer masking coincides with the peak in readout attention to question tokens around layer 14 (Figure~\ref{fig:readout_targets}). This identifies middle-layer distractor-to-text attention as a promising target for distractor invariance.

\section{Mitigating Distraction Through Distractor Invariance}

As established in Section~\ref{sec:knockouts}, middle-layer distractor-to-text attention serves as a key pathway for in-context corruption. While blocking this pathway mitigates errors, hard intervention relies on distractor labels at test time. To bypass this constraint, we train the model to output consistent answer distributions regardless of distractor presence, minimizing output divergence between clean and distractor-augmented contexts during fine-tuning.

\paragraph{Distractor Invariance Objective.} 
For an aligned group of inputs sharing the same multiple-choice question and target image, let $p_N$ denote the model's probability distribution over the answer options at distractor count $N$. We augment the standard cross-entropy loss with a Kullback-Leibler (KL) divergence penalty relative to the clean baseline ($N = 0$ distractors):

\vspace{-0.4cm}

\begin{equation}
\mathcal{L} \;=\; \mathcal{L}_{\text{CE}} \;+\;
\frac{\lambda}{|\mathcal{N}|}\sum_{N \in \mathcal{N}}
\mathrm{KL}\!\left(\operatorname{sg}\big[p_{0}\big] \,\big\|\, p_{N}\right),
\qquad \mathcal{N} = \{1,2,4,8\}.
\end{equation}

\vspace{-0.2cm}

The divergence is computed over the complete option distribution rather than $P(\text{correct})$ alone, constraining how probability mass is reallocated among alternative options under distraction. The clean distribution $p_0$ is produced by the model itself within the same training step and is detached via a stop-gradient operator. We evaluate regularization strengths $\lambda \in \{0.1, 0.5, 1.0, 2.0\}$.

\paragraph{Experimental Setup.} \label{sec:mitigation-setup} 
We fine-tune \textit{Qwen2.5-VL-7B-Instruct} using LoRA (rank $r = 16$, $\alpha = 32$) on a subset of the FineVision \citep{wiedmann2025finevision} dataset. Each question is expanded to include distractor counts $N \in \{0, 1, 2, 4, 8\}$, yielding $12{,}708$ training and $1{,}000$ validation aligned groups. 

To isolate the impact of the consistency loss, we establish three primary baselines:
\vspace{-0.1cm}
\begin{enumerate}[leftmargin=*, topsep=2pt, itemsep=0pt, parsep=0pt, partopsep=0pt]
\item \textbf{CE-Only Control:} Fine-tuned on identical data and configuration, but with $\lambda = 0$, isolating the effect of distribution alignment beyond data augmentation.
    \item \textbf{Inter-Image Attention Masking:} Re-implementation of \citet{das2026more}, restricting image tokens in deep decoder layers ($14$--$27$) to their local image blocks during training and inference.
    \item \textbf{Delimiter Token Scaling (DTS):} Training-free method by \citet{lee2026enhancing}, which rescales image-delimiter hidden states by $s = 8$ across layers $0$--$3$ on the base model.
\end{enumerate}
Full hyperparameter details are available in Appendix~\ref{app:experimental_setup_finetuning}. We evaluate models across our five held-out benchmarks at $N \in \{0,1,2,3,5\}$ distractors, reporting circular accuracy.

\paragraph{Results.} Table~\ref{tab:setB_v3_circular_main} reports circular accuracy across all benchmarks and distractor levels. The KL-regularized model ($\lambda = 1.0$) achieves the highest pooled mean performance under distracted conditions at $69.6\%$, outperforming both the CE-only control ($66.6\%$) and the base model ($59.3\%$). At five distractors, $\lambda = 1.0$ maintains $67.0\%$ accuracy, compared to $64.1\%$ for CE-only and $51.4\%$ for the base model. Performance improvements over the CE control are more pronounced on CLEVR-size, MIMIC, and What'sUp (Table~\ref{tab:setB_v3_circular}). Increasing the regularization weight beyond optimal levels ($\lambda = 2.0$) yields a slight decrease in pooled performance ($67.8\%$). For comparison, the reference baselines achieve pooled means of $67.0\%$ (Inter-Image Attention Masking) and $61.0\%$ (DTS).

These gains persist when the target is named: under \textit{Ground-Truth} and \textit{Two-Step} prompting, KL-clean $\lambda = 1.0$ improves circular accuracy by 15.4-16.8 points at every distractor count (Table~\ref{tab:kl_prompts_v3}), so training reduces in-context corruption, not just selection errors.

\input{tables/tab_setB_v3_circular_main_ctrl}

\begin{figure}[t]
    \centering
    \includegraphics[width=\linewidth]{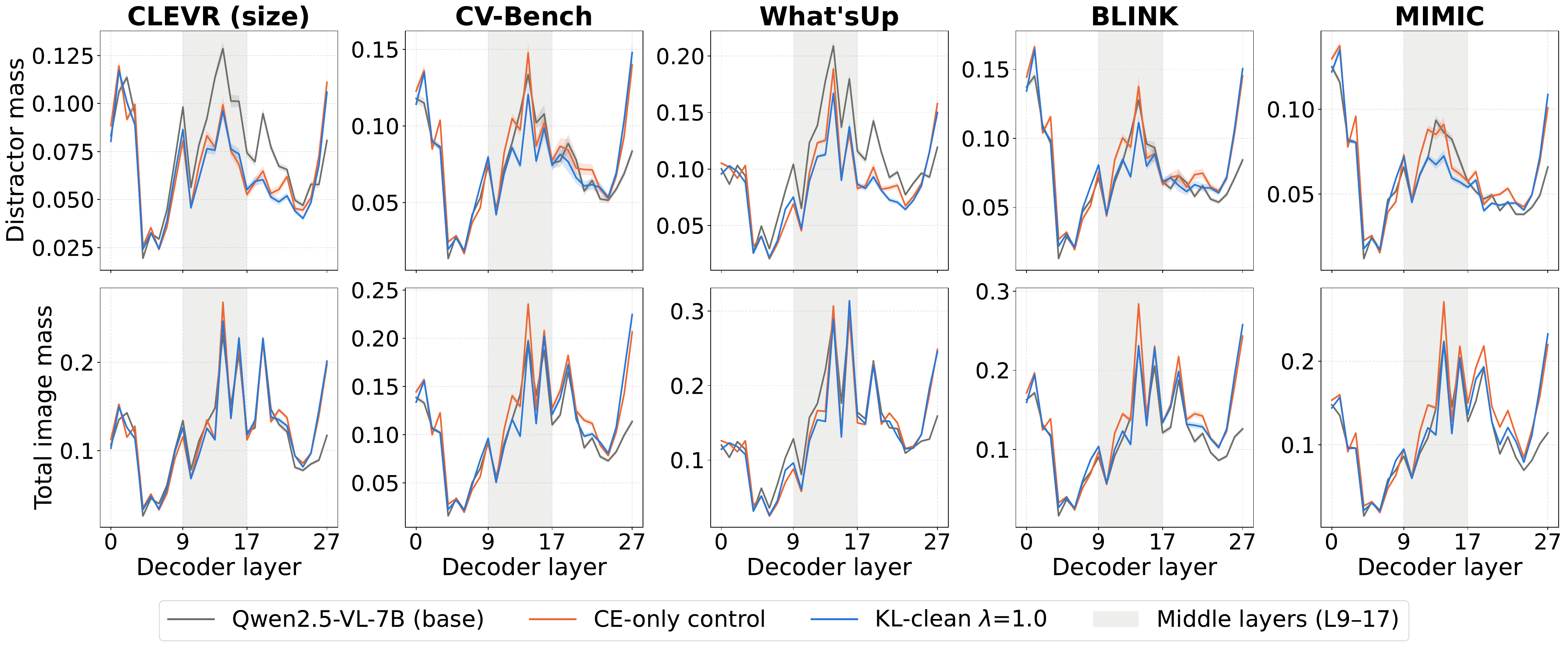}
    \vspace{-0.8cm}
\caption{\textbf{Question-token attention to images by decoder layer.} Attention is averaged over heads and question-token query positions, then summed over each image's visual-token span. Rows show distractor attention mass, and total image attention mass. Columns show the five benchmarks at $k=5$ ($n=150$ per benchmark; $n=300$ for What'sUp). Curves compare the frozen base model, the CE-only control, and KL-clean model $\lambda = 1$.}
    \label{fig:attention_after_training}
  \vspace{-0.5cm}
\end{figure}

\paragraph{Attention after fine-tuning.} Although the objective does not constrain attention, fine-tuning reshapes the pathway identified in Section~\ref{sec:knockouts} (Figure~\ref{fig:attention_after_training}). Total image attention remains stable, but question tokens shift their image attention from distractors toward the target, concentrated in middle layers 9–17, where distractor-to-text knockouts were most effective. The shift is largest on CLEVR-size and What'sUp, which also show the largest accuracy gains. It appears in both CE-only and KL-clean models, indicating that it stems from training on distractor-augmented data, while the consistency term adds further accuracy gains (Table~\ref{tab:setB_v3_circular_main}).

\section{Conclusion}
Irrelevant images can degrade VLM performance even when target visual context is correctly identified. To systematically study this failure mode, we adapt five visual perception benchmarks to incorporate controlled distractor images. Attention knockout experiments on Qwen2.5-VL reveal that a substantial share of visual interference flows through text-token representations; specifically, blocking distractor-to-text attention in layers $9$--$17$ restores $47\%$ of failure cases caused by distractors. To mitigate this vulnerability, we enforce answer-level distractor invariance by aligning model output distributions on clean and distracted inputs. Across five held-out benchmarks, optimizing KL divergence to clean distributions ($\lambda = 1.0$) improves pooled circular accuracy by 16 percentage points over the base model under five-distractor conditions. Attention analysis further indicates that fine-tuning shifts question-token attention toward target visual features, particularly within intermediate layers. Finally, our evaluations rely on randomly sampled distractors, leaving open how specific distractor properties impact robustness. Investigating structured spatial distractors will be essential to decouple perceptual confusion from spatial reasoning failures, while extending this framework to video domains will test robustness under dynamic, time-varying distractor contexts.

\bibliography{preprint}
\bibliographystyle{preprint}

\appendix
\section{Appendix}

\subsection{Experimental Set-Up}

\subsubsection{Dataset Augmentation}
\label{app:augmentation}

We will release all distractor-augmented benchmarks upon acceptance to support reproducibility.

We study distraction in a single-target-image setting: each question concerns one relevant image, presented either alone or alongside \(k\) distractor images that do not support the answer. The underlying task and its ground-truth answer are preserved, but the surface form is adapted where a single-image phrasing would no longer be well posed once several images are shown. Questions lose their references to a specific image, so that \textit{How many organs are in the image?} becomes \textit{How many organs are there?}; the rewritten question is used at every level, including $k{=}0$, so wording never covaries with the number of images. Answer options are adapted in the same spirit where the original task was framed around one image: What'sUp, for instance, is natively a caption-selection task whose options are complete sentences, which is ambiguous when a caption may also hold a distractor. We reframe it as a question over spatial relations, with the relation alone as the option. We build $k \in \{1,2,3,5\}$ for every benchmark, together with the matched single-image condition (\(k=0\)), using the same rewritten question and answer options at every distractor level, and hold the row set fixed across levels so that a degradation curve compares the same questions at every point.

All benchmarks except MIMIC share one skeleton: (1) build a distractor candidate pool, (2) exclude images that could legitimately answer the question, (3) vet what remains, and (4) place eligible distractors alongside the relevant image in the shuffled order. MIMIC skips steps 1–3 because each item already comes with its own distractors. Table~\ref{tab:pipeline_steps} gives the per-benchmark steps. 

\input{tables/tab_pipeline_steps}

\paragraph{Distractor pools.} Table~\ref{tab:pool_sources} details the source and size of each distractor pool. Where available, distractors are drawn directly from the benchmark's own image distribution to ensure they are visually confusable with target images and cannot be dismissed based on rendering style alone. BLINK starts from this: because its images contain drawn annotations essential to the questions and span multiple source corpora, it uses a "matched-domain" pool of real photographs matched to image kind rather than exact distribution. CV-Bench uses a hybrid strategy: queries whose objects fall within COCO's categories draw from COCO, while the rest fall back to the benchmark's own box-free images. Both CV-Bench and BLINK draw their COCO images from the test2017 split. Because that split publishes no instance annotations, both routes are vetted by the answerability gate rather than by ground-truth absence. MIMIC does not need a candidate pool: each item natively provides six distractor images, and we build each distraction level by subsampling $k \in {1, 2, 3, 5}$ of these native distractors and shuffling them with the relevant image.

\input{tables/tab_pool_sources}

\paragraph{Vetting.} A candidate image is only a valid distractor if it cannot answer the target question. Because metadata formats vary, we apply the strongest guarantee supported by each benchmark rather than a single uniform rule (Table~\ref{tab:vlm_gates}). Only CLEVR permits exact verification, since its scene graphs allow ground-truth attribute matching. MIMIC requires no additional vetting, since we reuse the distractors provided by the benchmark itself. Where no such guarantee exists, we ask the row's own question of each candidate \emph{in isolation} and reject the whole set if any image could support an answer; verdicts are single-token and strictly parsed, and a rejected set is redrawn, up to $15$ attempts. 

\begin{figure}[h]
\begin{lstlisting}[style=prompt]
Question: Where is the beer bottle?
Answer choices: Under an armchair, On a armchair, To the left of a armchair, To the right of a armchair

Does this image contain any of the objects the question asks about, or could this image on its own support answering the question?

Answer with exactly one word: YES if it contains any of those objects or could support an answer, NO if it is completely unrelated. No explanation.
\end{lstlisting}
\caption{\textbf{Answerability gate prompt.}}
\label{fig:answerability}
\end{figure}

\input{tables/tab_vlm_gates}

\paragraph{Transformations.} Accepted distractors are resized to match the reference image's exact dimensions (Table~\ref{tab:transformations}). Beyond resizing, only CV-Bench requires active visual editing. In CV-Bench, $1{,}619$ of $2{,}560$ questions target an object highlighted by a colored box drawn on the image; unedited distractors lacking these box strokes would allow models to trivially identify the target image via a "pick the image with markers" heuristic without reading the question. We address this by drawing synthetic boxes on distractors using the exact colors named in the question. Marker pixel area is matched to that of the target image with added per-distractor jitter, placing the target image's marker density within the overall distribution rather than at an extreme. 

\input{tables/tab_transformations}

\subsubsection{Examples}

\begin{figure}[h]
    \centering
    \includegraphics[width=\linewidth]{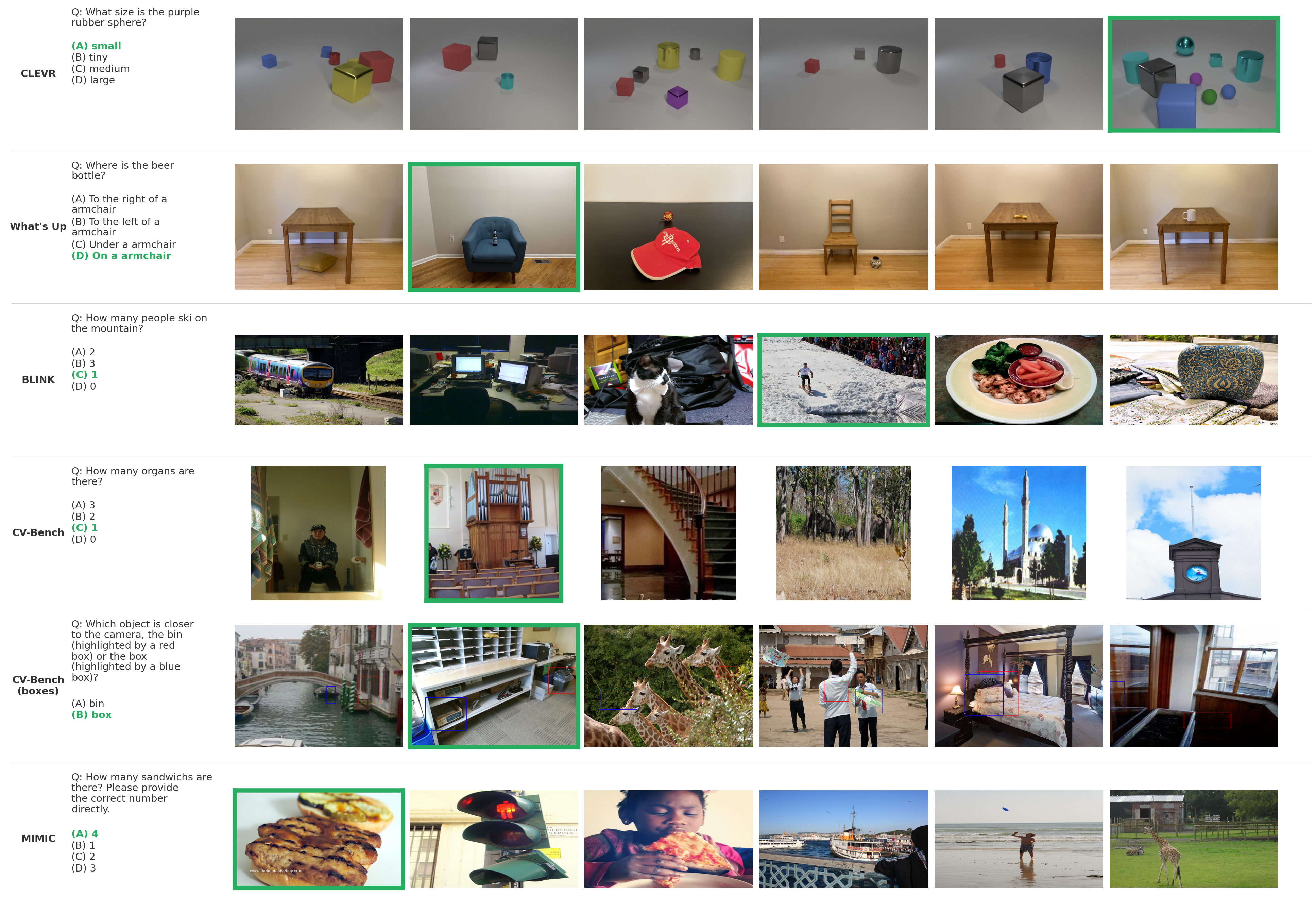}
    \vspace{-0.5cm}
    \caption{\textbf{Example items at the maximum distraction level ($k=5$).} Each row shows a question, its choices (correct answer in green), and the six images shown to the model: one relevant image (green outline) among five distractors from the same image distribution, in shuffled order.}
    \label{fig:examples}
\end{figure}

Figure~\ref{fig:examples} presents sample benchmark items at the highest distraction level ($k=5$). Each item consists of one relevant target image and five distractors sampled from the same domain, presented in randomized order alongside the question and options. All prompts use neutral phrasing stripped of single-image references so that question text does not vary with $k$. For CV-Bench's annotated questions (row 5), distractors feature matched synthetic boxes to ensure marker presence cannot be exploited as a shortcut.

\subsubsection{Answer Distribution and Relevant Image Distribution}

\begin{figure*}[h]
    \includegraphics[width=\linewidth]{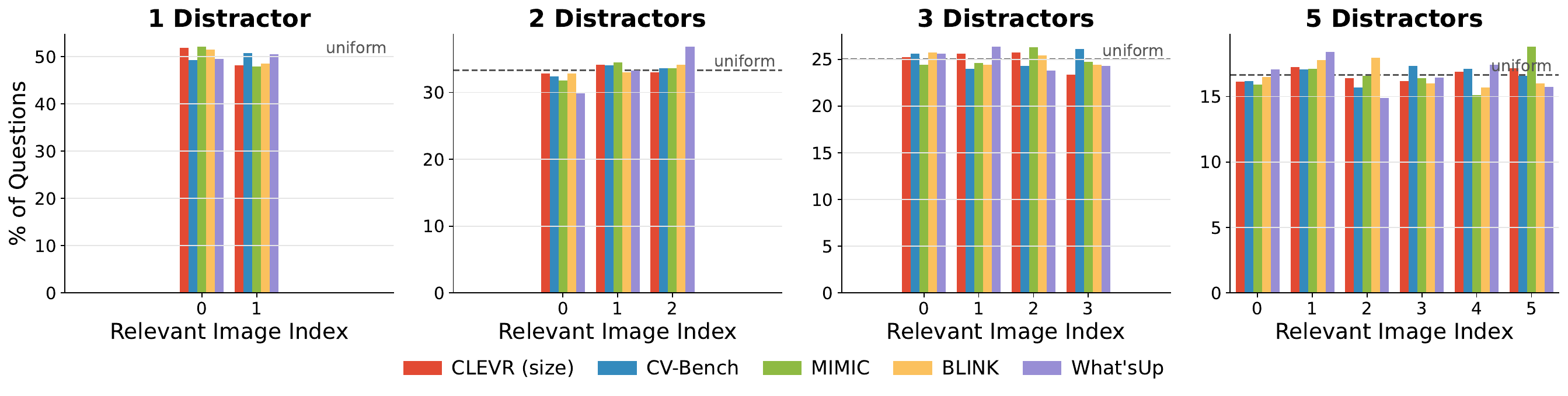}
    \vspace{-0.5cm}
    \caption{\textbf{The relevant image is placed uniformly across positions at every distractor level.} For each benchmark, we plot the fraction of questions in which the question-relevant image occupies each slot, split by the number of distractors (1, 2, 3, 5). Distributions track the uniform reference (dashed line) across all positions and benchmarks, confirming that a model cannot exploit a fixed-position prior to locate the relevant image.}
    \label{fig:answer_distribution}
\end{figure*}

Because our distractor construction inserts the relevant image at a random slot, its position carries no information about the answer. Figure~\ref{fig:answer_distribution} verifies this: across all benchmarks and distractor counts, the relevant-image index is close to uniform, so accuracy under distraction reflects genuine cross-image grounding rather than a positional shortcut.

The same holds for the answers themselves: because each question is evaluated under all four circular rotations of its options, the correct answer appears in every position exactly once, eliminating answer-position bias by construction.

\subsection{Inference Details}
\label{sec:inference_details}
Before examining whether selection alone explains distractor degradation, we must ensure that observed accuracy drops reflect genuine interference rather than context-truncation artifacts.

\subsubsection{Budget Selection}

Multi-image evaluation with $k=5$ distractors requires careful management of the context length. Each image consumes visual tokens proportional to its resolution, and with 6 images per sample (1 relevant + 5 distractors), an uncapped budget leaves the input length unbounded and dependent on each benchmark's native resolution. We therefore fix a budget, and verify that doing so costs no accuracy.

\begin{figure}[h]
    \centering
    \includegraphics[width=0.7\linewidth]{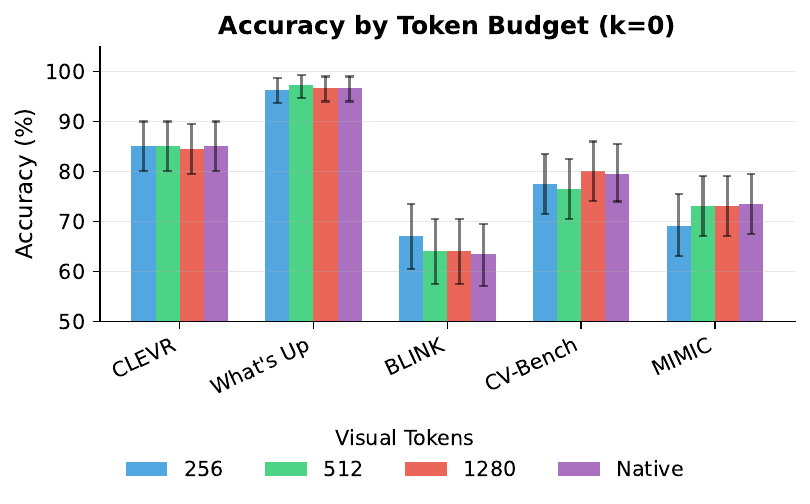}
    \vspace{-0.5cm}
    \caption{\textbf{Token budget selection for multi-image evaluation.} Accuracy at $k{=}0$ (single image) across token budgets; 1280 tokens matches native resolution.}
    \label{fig:budget_selection}
\end{figure}

We evaluated four visual token budgets (256, 512, 1280, and native (uncapped)) on 200 samples per benchmark (400 for What'sUp) at $k=0$ (single image) using Qwen2.5-VL-7B. Figure~\ref{fig:budget_selection} shows that 1280 tokens matches native resolution closely, with every benchmark within $0.5$ points; the smaller budgets diverge more (up to $4.5$ points on MIMIC at 256).

Based on these results, we use 1280 visual tokens per image (corresponding to 1M pixels) for all multi-image experiments, ensuring no context truncation while maintaining accuracy parity with native resolution.

\subsubsection{Prompts}

\paragraph{Regular.}
No hint about distractors; the model must implicitly identify the relevant image.

\begin{lstlisting}[style=prompt]
{question}
A. {choice_a}  B. {choice_b}
C. {choice_c}  D. {choice_d}
Answer with the option's letter from the given choices directly.
\end{lstlisting}

\paragraph{Ground Truth (Oracle).}
The relevant image index is revealed to the model (upper bound).

\begin{lstlisting}[style=prompt]
You are given {n} images (Image 1, ..., Image {n}).
Only Image {k} is relevant for answering the question; ignore the other images.
{question}
A. {choice_a}  B. {choice_b}
C. {choice_c}  D. {choice_d}
Answer with exactly one letter (A, B, C, D). No punctuation, no explanation.
\end{lstlisting}

\paragraph{Two-Step/Hard Isolation, Phase 1: Image Selection.}
First, ask only which image is relevant.

\begin{lstlisting}[style=prompt]
You are given {n} images. Only one is relevant for answering the question below.
{question}

Which image should be used to answer the question?

A. First image
B. Second image
...
Answer **only** with A, B, ... No additional text.
\end{lstlisting}

\paragraph{Two-Step, Phase 2: Answer.}
Condition the answer on the selected image from Phase 1.

\begin{lstlisting}[style=prompt]
You are given {n} images (Image 1, ..., Image {n}).
Only Image {selected} is relevant for answering the question; ignore the other images.
{question}
A. {choice_a}  B. {choice_b}
C. {choice_c}  D. {choice_d}
Answer with exactly one letter (A, B, C, D). No punctuation, no explanation.
\end{lstlisting}

\paragraph{Hard Isolation, Phase 2: Answer.}
After Phase 1, distractors are \emph{physically removed} from context; only the selected image is shown with the Regular prompt.

\begin{lstlisting}[style=prompt]
{question}
A. {choice_a}  B. {choice_b}
C. {choice_c}  D. {choice_d}
Answer with the option's letter from the given choices directly.
\end{lstlisting}

\noindent\textit{Note:} Two-Step Phase 2 and Ground Truth use the same answer template; they differ in whether the relevant-image index is predicted or supplied.

\subsection{Results}

\subsubsection{Baseline}

Table~\ref{tab:baseline_v3_rnd} reports the per-benchmark accuracies underlying Figure~\ref{fig:degradation}, for distractor counts $k \in \{0,1,2,3,5\}$. From zero to five distractors, mean-rotation accuracy declines on every benchmark for all three models. What'sUp shows the largest drops: $23.0$, $26.4$, and $31.7$ percentage points for Qwen2.5-VL, InternVL3, and LLaVA-OV, respectively.  The smallest drops are on BLINK ($6.0$ to $8.0$ points across the three models) and on MIMIC, which is the most resilient benchmark for Qwen2.5-VL ($5.3$) and LLaVA-OV ($6.2$).

\input{tables/tab_baseline_rnd}

\subsubsection{Prompting Strategies on Qwen2.5-VL}
\label{app:prompting_strategies_qwen}
Table~\ref{tab:prompting_v3_qwen_rnd} reports the full results with the different prompts on Qwen2.5-VL.  

\input{tables/tab_prompting_qwen_rnd}

\subsubsection{Prompting Strategies on InternVL3 and LLaVA-OV}
\label{app:prompting_strategies}

\begin{figure*}[h]
    \centering
    \includegraphics[width=\linewidth]{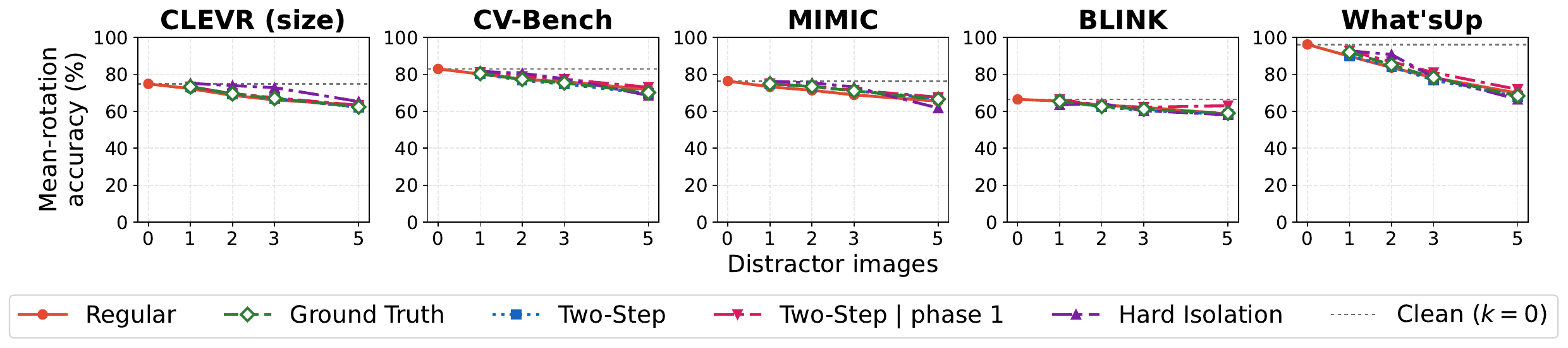}
    \vspace{-0.5cm}
     \caption{\textbf{Prompting strategies under distraction (InternVL3-8B)}. Mean-rotation accuracy as a function of the number of distractor images for four inference strategies (Regular, Ground-truth, Two-step, and Hard isolation) on each benchmark.}
    \label{fig:prompts_internvl3}
\end{figure*}

\begin{figure*}[h]
    \centering
    \includegraphics[width=\linewidth]{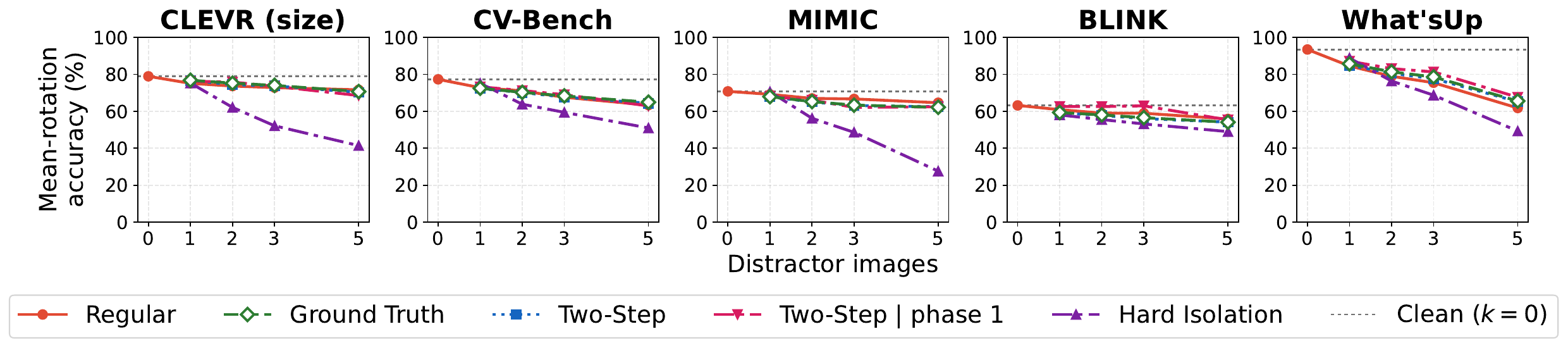}
      \vspace{-0.5cm}
     \caption{\textbf{Prompting strategies under distraction (LLaVA-OV)}. Mean-rotation accuracy as a function of the number of distractor images for four inference strategies (Regular, Ground-truth, Two-step, and Hard isolation) on each benchmark.}
    \label{fig:prompts_llava}
\end{figure*}

Figures~\ref{fig:prompts_internvl3} and~\ref{fig:prompts_llava} replicate the prompting-strategy analysis of Fig.~\ref{fig:strategies} on InternVL3-8B and LLaVA-OneVision. Two model-specific patterns emerge.

On InternVL3, the ground-truth, two-step, and hard-isolation curves are essentially indistinguishable from the regular baseline across all five benchmarks: revealing the relevant image, or even removing the distractors entirely, yields no consistent gain. Conditioning instead on correct phase-1 selection separates the two failures: hard isolation then recovers clean-image accuracy, whereas ground-truth and two-step do not. The residual end-to-end penalty is thus attributable to selection errors compounding with distractor interference, and the pattern mirrors the main-text finding: locating the relevant image is not sufficient—only removing the distractors from the input restores performance.

\begin{figure*}[h]
  \centering
  \includegraphics[width=\textwidth]{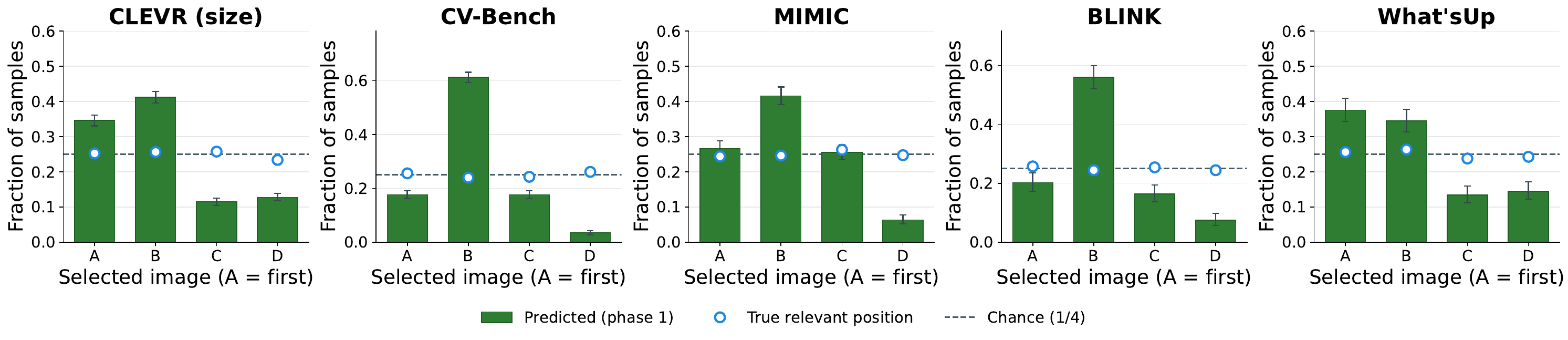}
  \vspace{-0.5cm}
  \caption{\textbf{Phase-1 image selection is positionally biased toward option B (second image) in LLaVA-OV.} We zoom on the case where there are 3 distractors. For each benchmark, green bars give the distribution over the image slot selected in phase 1 (A${=}$first image, ...). Error bars are $95\%$ Wilson score confidence intervals. Although the relevant image is near-uniformly distributed across positions, the model selects option B far above chance ($25\%$) in every benchmark.}
\label{fig:phase1_positional_bias_llava_ov_twostep_4imgs_circular}
\end{figure*}

LLaVA-OV illustrates the other half of the decomposition: whereas InternVL3 selects reliably and suffers mainly from residual interference, LLaVA-OV's dominant failure is selection itself. Hard isolation (the best strategy for Qwen2.5-VL) is the worst here, falling well below the regular baseline as distractors are added. Because the two-step and hard-isolation curves are reported end-to-end, they inherit phase-1 selection errors, and LLaVA-OV's selection accuracy collapses toward chance with more distractors (Tab.~\ref{tab:image_selection_v3_twostep_rnd}). Hard isolation is particularly sensitive to selection errors: choosing the wrong image discards the relevant image and removes the visual evidence needed to answer reliably. Inspecting the phase-1 responses (Figure~\ref{fig:phase1_positional_bias_llava_ov_twostep_4imgs_circular}) reveals the source of the collapse: the model exhibits a strong positional bias toward option B. It selects the second image in 35-60\% of cases across all five benchmarks, well above the $25\%$ chance rate (one-proportion binomial test, $p<10^{-10}$), even though the relevant image is near-uniformly distributed over positions.

\begin{figure*}[h]
   \centering
    \includegraphics[width=\linewidth]{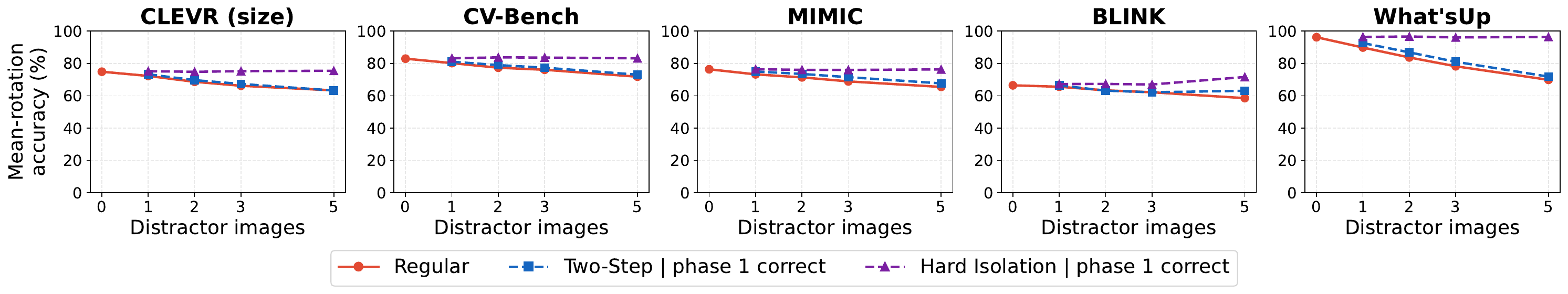}
      \vspace{-0.5cm}
    \caption{\textbf{Correctly locating the relevant image does not undo distractor-induced degradation. (InternVL3-8B)}}
    \label{fig:internvl3_found} 
\end{figure*}

\begin{figure*}[h]
   \centering
    \includegraphics[width=\linewidth]{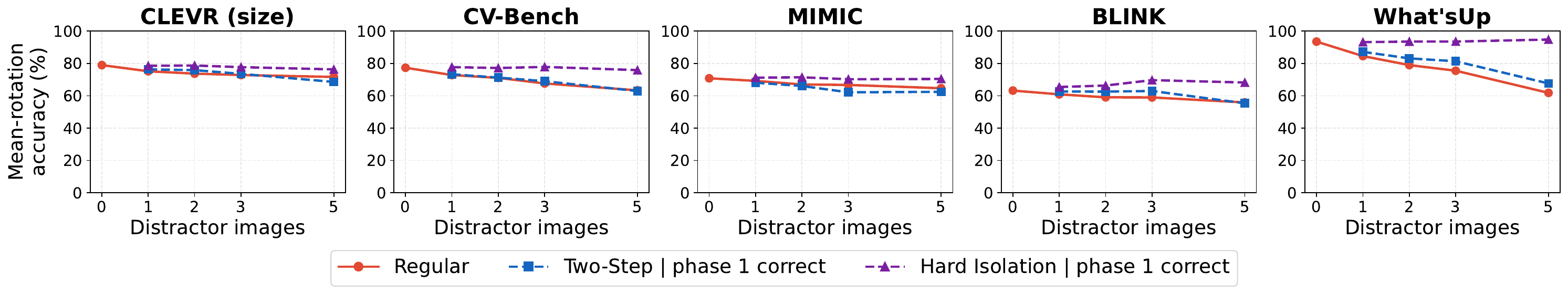}
      \vspace{-0.5cm}
    \caption{\textbf{Correctly locating the relevant image does not undo distractor-induced degradation. (LLaVA-OneVision-7B)}}
    \label{fig:llava_found} 
\end{figure*}

\subsection{Sequence Length}
 A natural hypothesis for distractor-induced degradation is that longer visual context simply dilutes attention or exceeds the model's effective context window. To test this, we construct a \textit{black-image control} for every distractor variant: each distractor image is replaced by a solid black image of identical pixel dimensions, preserving the number and positions of visual tokens while removing distractor \textit{content}.

\begin{figure*}[h]
     \centering
     \includegraphics[width=\linewidth]{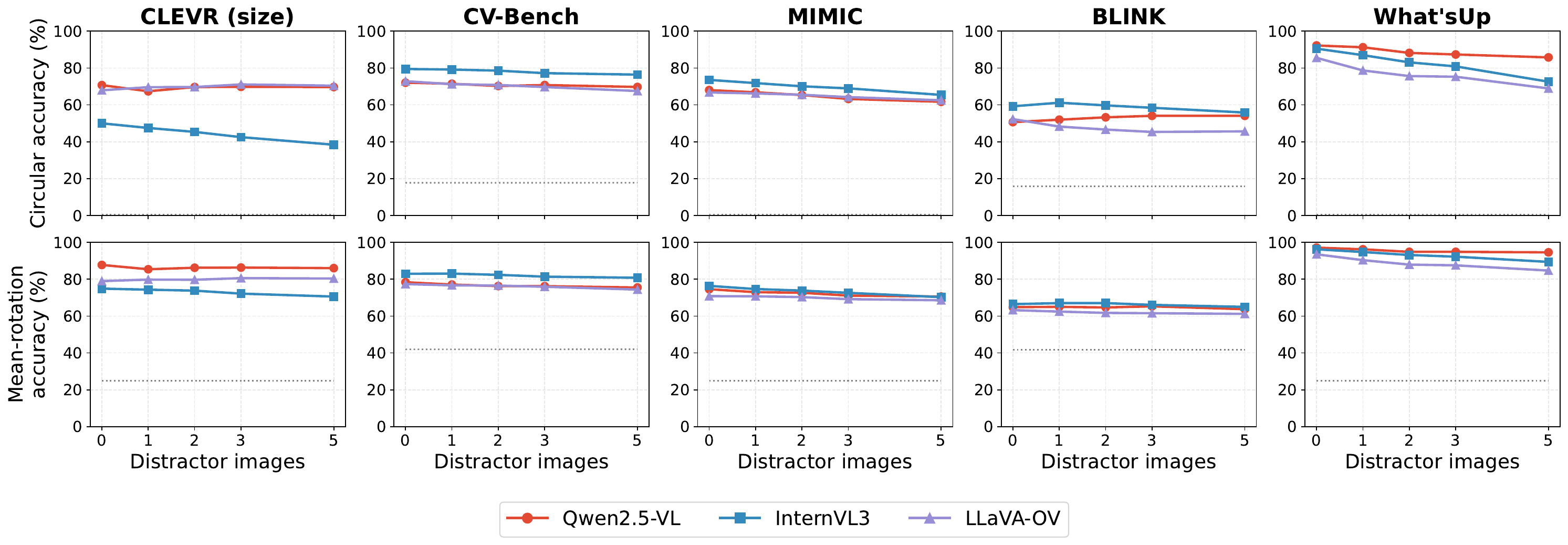}
     \vspace{-0.5cm}
     \caption{\textbf{Degradation under black-image distractors (sequence-length control).} Same layout as Figure~\ref{fig:degradation}, but each distractor is replaced by a solid black image of identical dimensions. The curves are flatter, demonstrating that sequence length alone does not explain the distractor penalty.}
     \label{fig:degradation_black}
 \end{figure*}

 \paragraph{Results.}
 Figure~\ref{fig:degradation_black} shows accuracy as a function of the number of distractors under the black-image control. In stark contrast to the random distractors (Figure~\ref{fig:degradation}), the curves are more robust: averaged over three models and five benchmarks, mean-rotation accuracy drops by only 0.9 points at $k{=}1$ and 3.2 points at $k{=}5$, compared to 3.7 and 13.3 points respectively under random distractors.

 Table~\ref{tab:content_fraction_v3} quantifies the per-benchmark degradation. We report the fraction of the penalty attributable to visual content (as opposed to sequence length):
 \[ \text{Content fraction} = \frac{\Delta_{\text{rnd}} - \Delta_{\text{black}}}{\Delta_{\text{rnd}}}
 \]
 where $\Delta$ denotes the drop from the clean baseline $k{=}0$ to $k{=}5$. Averaged across benchmarks, the content fraction ranges from 69\% to 81\% across the three models.

 \input{tables/tab_content_fraction}

Benchmarks differ in how much degradation persists under the matched-length black-image control. On CLEVR (size), What'sUp, and BLINK, replacing real distractors with black images substantially reduces the accuracy loss, suggesting that visual content contributes beyond sequence length alone. On BLINK, this reduction accounts for $72$-$82\%$ of the original penalty. MIMIC retains a larger proportion of the degradation under the control: for Qwen2.5-VL, approximately three-quarters of the penalty persists with black images. This pattern is consistent with a greater contribution from sequence length on MIMIC.

\subsection{Where Does the Interference Occur? An Internal Representation Analysis}

\subsubsection{Attention Analysis}
\label{app:attention_analysis}

All attention measurements use Qwen2.5-VL-7B-Instruct ($28$ decoder layers, $28$ heads, indexed $0$-$27$) at \texttt{max\_pixels}$=1,003,520$, the budget used for the accuracy evaluations. Values are head-averaged and normalized by the query row's total, so each row sums to $1$ over the token groups and
``group $g$ receives $x\%$ of the attention'' is well defined.

\paragraph{Token groups.} Every sequence position is assigned to one of the groups illustrated in Table~\ref{tab:token_groups}. 

\begin{table}[h]
  \centering\small
  \caption{\textbf{Token groups}.}
  \begin{tabular}{ll}
    \toprule
    \textsc{group} & \textsc{contents} \\
    \midrule
    system / header     & attention sink (position $0$) + system prompt + user header \\
    target image        & visual tokens of the relevant image \\
    distractor images   & visual tokens of each distractor, in positional order \\
    question            & question stem + answer choices (content words only) \\
    readout (answer)    & the last position, where the answer is produced \\
    separator           & vision delimiters (between the image blocks), answer-format \\
                        & instruction and generation scaffold (after the last image) \\
    \bottomrule
  \end{tabular}
  \label{tab:token_groups}
\end{table}

\begin{figure}[h]
    \centering
    \includegraphics[width=\linewidth]{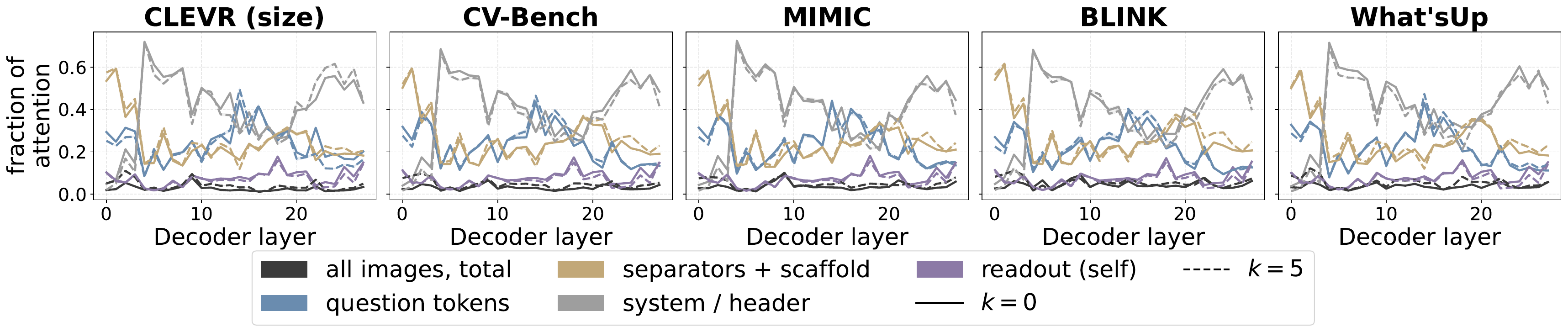}
    \caption{\textbf{Readout-token attention across decoder layers in Qwen2.5-VL-7B-Instruct.} Curves show the fraction of attention allocated to all image tokens (black), question tokens (blue), separators and prompt-scaffold tokens (gold), system and header tokens (gray), and the readout position itself (purple). Solid lines denote clean inputs ($k{=}0$), and dashed lines denote inputs with five distractor images ($k{=}5$). Results use $n{=}150$ examples per benchmark, except What'sUp ($n{=}300$).}
    \label{fig:readout_full}
\end{figure}

\paragraph{Where the readout's attention goes.} Figure~\ref{fig:readout_full} decomposes the readout row across all groups.

\subsubsection{Attention Knockouts}
\label{app:knockout}
We use attention knockouts to test whether specific information-flow routes contribute to distraction-induced errors. Unlike the descriptive inference analysis in Section~\ref{sec:evaluation}, these interventions modify the model's computation at inference time. For selected query-key pairs, we set the corresponding entry of the additive attention mask to $-\infty$ before the softmax. The masked query therefore cannot attend to the selected key or receive information through its value vector. Model parameters and all other attention connections remain unchanged.

Table~\ref{tab:knockout_spans} summarizes the interventions evaluated in our experiments. \textbf{D} blocks the distractors' visual tokens from every position after the final image (question, answer-format instruction, generation scaffold and readout). This excludes the system header, positioned before the images, and the delimiters interleaved between earlier image blocks. \textbf{R} is the readout alone and \textbf{P} is every position that is not itself a distractor token, so $\mathbf{R} \subset \mathbf{D} \subset \mathbf{P}$.  \textbf{MM} sits outside this chain: it prevents each \emph{image} from reading what precedes it (i.e., the system header and all earlier images) and it does so for every image, the relevant one included.

\input{tables/tab_knockout_spans}

Table~\ref{tab:knockout-ci} reports the middle-layer attention-knockout results separately for each benchmark. We apply each intervention in layers 9-17 of Qwen2.5-VL-7B-Instruct with five distractor images. A \emph{broken} item is answered correctly in the corresponding clean condition but incorrectly when distractors are added; a \emph{right} item remains correct despite the distractors. For each intervention, \textsc{Recov./Broken} measures the fraction of broken items whose correct answer is restored, whereas \textsc{Broken/Right} measures the fraction of previously correct items that the intervention causes the model to answer incorrectly. Confidence intervals are percentile 95\% bootstrap intervals obtained from 10,000 item-level resamples within each benchmark.

\input{tables/tab_knockout_ci_middle}

 Table~\ref{tab:knockout-overlap} shows that \textbf{D} and \textbf{P} have strong overlap: $92\%$ of items repaired by \textbf{D} are also repaired by \textbf{P}. Table~\ref{tab:knockout-bands} reports recovery for every condition across the four layer bands.
 
\input{tables/tab_knockout_overlap_middle}

\input{tables/tab_knockout_bands}

\subsection{Mitigating Distraction Through Distractor Invariance}

\subsubsection{Experimental Setup}
\label{app:experimental_setup_finetuning}

\paragraph{Model and adaptation.} All fine-tuned models start from Qwen2.5-VL-7B-Instruct. We train LoRA adapters with rank $16$, $\alpha = 32$, and dropout $0.05$. The vision encoder and the vision-language merger are frozen, so every intervention we report acts on the language model's use of visual evidence rather than on the visual features themselves.

\paragraph{Optimization.} We use AdamW at a learning rate of $2 \times 10^{-4}$ with a cosine schedule. All arms share the same data order and LoRA configuration, so differences between arms are attributable to the objective alone.

\paragraph{Validation, early stopping, and checkpoint selection.} We hold out a fixed random subset of $1{,}000$ validation questions (Appendix~\ref{app:training_set_construction}) and evaluate at the end of each epoch on all five variants ($k \in \{0,1,2,4,8\}$). Early stopping uses a patience of $2$ epochs, and the exported checkpoint maximizes mean accuracy over the five variants.

\subsection{Training Set Construction}
\label{app:training_set_construction}

\paragraph{Source and scope.} The mitigation experiments train on a distractor ladder built from FineVision \citep{wiedmann2025finevision}, deliberately drawn from a different domain than the perception benchmarks we evaluate on. Training rows come from thirteen FineVision sub-datasets grouped into five categories: general scene VQA (A-OKVQA), general VQA (Visual7W), science and diagram reasoning (AI2D, ScienceQA, ScienceQA without context), medical imaging (PMC-VQA), and math and geometry (GeoQA+, Geometry3K, Geo170K, InterGPS, UniGeo, MAVIS-metagen, MAVIS-rule). We keep only multiple-choice turns. The source pool includes $12{,}708$ training questions and $1{,}000$ validation questions, composed of A-OKVQA $33\%$, science and diagram reasoning $19\%$, PMC-VQA $19\%$, Visual7W $16\%$, and math and geometry $13\%$. None of the five evaluation benchmarks contributes to training. Hence, gains reported in Section~\ref{sec:mitigation-setup} measure transfer of a distraction-robustness behavior rather than in-domain adaptation.

\paragraph{Distractor construction.} Distractors are sampled from the same sub-dataset as the source row (e.g., geometry questions receive geometry distractors). Domain-matching prevents models from shortcutting the task using low-level visual differences, forcing them to resolve the question-image relation. Following the LoCoVQA pipeline~\citep{sharma2024losing}, we apply a two-stage VLM filter to each domain-matched candidate, which is more likely than a randomly sampled image to contain question-relevant content. The first stage extracts the entities present in the candidate image, and the second returns a binary verdict indicating whether any extracted entity is relevant to the target question. Failed candidates are resampled up to 15 times; rows that cannot be filled are dropped. Both filter prompts enforce a strict constraint on people: if the question involves a person, distractors must contain no people to avoid referential ambiguity.

\subsection{Clean-Anchored Answer Consistency}

Table~\ref{tab:setB_v3_circular} reports circular accuracy for every benchmark, distractor count, and training condition.

\input{tables/tab_setB_v3_circular_extra}

To examine whether clean-anchored answer consistency affects the attention pathways identified in our mechanistic analysis, we compare attention patterns under CE-only and CE+KL fine-tuning. We measure how question tokens allocate attention to target and distractor images across layers, focusing on the middle layers where distractor-to-text knockouts were most effective. This analysis tests whether regularizing answer distributions also changes attention allocation, even without imposing an explicit attention constraint.

Finally, we ask whether these gains persist when the relevant image is identified explicitly rather than inferred. We evaluate the KL-clean model ($\lambda{=}1.0$) under the ground-truth and two-step prompts against the untrained base model (Table~\ref{tab:kl_prompts_v3}). The training transfers: circular accuracy improves by $15.4$-$16.8$ points at every distractor count under both prompts.

\input{tables/tab_kl_prompts_v3}

\end{document}

%% file: math_commands.tex
\usepackage{amsmath,amsfonts,bm}

\def\eqref#1{equation~\ref{#1}}

\def\1{\bm{1}}

\DeclareMathAlphabet{\mathsfit}{\encodingdefault}{\sfdefault}{m}{sl}
\SetMathAlphabet{\mathsfit}{bold}{\encodingdefault}{\sfdefault}{bx}{n}



%% file: tables/tab_image_selection_twostep_rnd.tex
\begin{table}[t]
\centering
\small
\caption{\textbf{Image-selection (phase-1) accuracy under the Two-Step protocol.} $k$ is the number of random distractors. Chance is $1/(k{+}1)$.}
\label{tab:image_selection_v3_twostep_rnd}
\resizebox{\textwidth}{!}{%
\begin{tabular}{l|rrrr|rrrr|rrrr}
\toprule
\textbf{Benchmark} & \multicolumn{4}{c|}{\textbf{Qwen2.5-VL}} & \multicolumn{4}{c|}{\textbf{InternVL3}} & \multicolumn{4}{c}{\textbf{LLaVA-OV}} \\
 & $k=1$ & $k=2$ & $k=3$ & $k=5$ & $k=1$ & $k=2$ & $k=3$ & $k=5$ & $k=1$ & $k=2$ & $k=3$ & $k=5$ \\
\midrule
CLEVR (size) & \cellcolor[RGB]{200,230,200}99.9 & \cellcolor[RGB]{200,230,200}99.7 & \cellcolor[RGB]{200,230,200}99.7 & \cellcolor[RGB]{200,230,200}99.4 & \cellcolor[RGB]{200,230,200}99.9 & \cellcolor[RGB]{201,230,199}98.6 & \cellcolor[RGB]{203,231,197}96.4 & \cellcolor[RGB]{217,234,187}84.1 & \cellcolor[RGB]{207,231,195}93.4 & \cellcolor[RGB]{237,240,171}66.0 & \cellcolor[RGB]{255,244,157}49.5 & \cellcolor[RGB]{255,228,178}29.8 \\
CV-Bench & \cellcolor[RGB]{204,231,197}96.3 & \cellcolor[RGB]{206,231,196}94.5 & \cellcolor[RGB]{207,232,194}92.9 & \cellcolor[RGB]{213,233,190}88.0 & \cellcolor[RGB]{206,231,195}94.1 & \cellcolor[RGB]{209,232,193}91.0 & \cellcolor[RGB]{217,234,187}84.5 & \cellcolor[RGB]{242,241,167}61.6 & \cellcolor[RGB]{208,232,193}92.0 & \cellcolor[RGB]{247,242,163}57.0 & \cellcolor[RGB]{255,242,160}46.2 & \cellcolor[RGB]{255,225,182}25.9 \\
MIMIC & \cellcolor[RGB]{200,230,200}99.7 & \cellcolor[RGB]{200,230,200}99.5 & \cellcolor[RGB]{200,230,200}99.2 & \cellcolor[RGB]{201,230,200}99.1 & \cellcolor[RGB]{200,230,200}99.9 & \cellcolor[RGB]{200,230,200}99.1 & \cellcolor[RGB]{204,231,197}96.3 & \cellcolor[RGB]{221,235,183}80.5 & \cellcolor[RGB]{201,230,200}99.0 & \cellcolor[RGB]{226,237,180}76.3 & \cellcolor[RGB]{236,239,172}67.1 & \cellcolor[RGB]{255,232,173}34.3 \\
BLINK & \cellcolor[RGB]{211,233,191}89.5 & \cellcolor[RGB]{215,234,188}85.6 & \cellcolor[RGB]{220,235,184}81.7 & \cellcolor[RGB]{220,235,184}81.7 & \cellcolor[RGB]{212,233,190}88.2 & \cellcolor[RGB]{217,234,186}83.8 & \cellcolor[RGB]{227,237,179}75.4 & \cellcolor[RGB]{252,244,158}51.9 & \cellcolor[RGB]{229,238,177}72.8 & \cellcolor[RGB]{250,243,160}53.9 & \cellcolor[RGB]{255,239,163}43.7 & \cellcolor[RGB]{255,226,181}26.5 \\
What'sUp & \cellcolor[RGB]{205,231,196}94.8 & \cellcolor[RGB]{208,232,194}92.2 & \cellcolor[RGB]{210,232,192}90.9 & \cellcolor[RGB]{216,234,188}85.4 & \cellcolor[RGB]{205,231,196}94.8 & \cellcolor[RGB]{209,232,193}91.7 & \cellcolor[RGB]{225,236,180}76.7 & \cellcolor[RGB]{246,242,163}57.7 & \cellcolor[RGB]{206,231,195}93.9 & \cellcolor[RGB]{228,237,177}73.7 & \cellcolor[RGB]{238,240,170}64.9 & \cellcolor[RGB]{255,232,173}34.3 \\
\bottomrule
\end{tabular}
}%
\end{table}

%% file: tables/tab_knockout_ci_middle_pooled.tex
\begin{table}[t]
  \centering
  \small
  \setlength{\tabcolsep}{5pt}
  \caption{\textbf{Knocking out distractor$\rightarrow$question attention repairs items that distraction broke.} Qwen2.5-VL, middle layers 9--17, pooled over 5 benchmarks at $k{=}5$ distractors. Recovery is the fraction of broken items knockout fixes; damage is the fraction of preserved items it damages. \textsc{range} and \textsc{macro} are the min--max and unweighted mean over individual benchmarks. Intervals are 95\% bootstrap CIs over items (10{,}000 resamples).}
  \label{tab:knockout-pooled}
  \begin{tabular}{llcccc}
    \toprule
    \textsc{cond.} & \textsc{route} & \textsc{recov./broken} & \textsc{range} & \textsc{macro} & \textsc{broken/right} \\
    \midrule
    P & IMG$_D$ $\rightarrow$ everything & 0.56 \tiny{[0.54, 0.58]} & 0.46--0.66 & 0.56 & 0.05 \tiny{[0.04, 0.05]} \\
    MM & all preceding $\rightarrow$ IMG (MIMIC) & 0.10 \tiny{[0.09, 0.12]} & 0.04--0.53 & 0.20 & 0.57 \tiny{[0.55, 0.58]} \\
    D & IMG$_D$ $\rightarrow$ all text & 0.47 \tiny{[0.45, 0.49]} & 0.44--0.54 & 0.49 & 0.05 \tiny{[0.04, 0.06]} \\
    R & IMG$_D$ $\rightarrow$ readout & 0.03 \tiny{[0.03, 0.04]} & 0.02--0.07 & 0.04 & 0.02 \tiny{[0.01, 0.02]} \\
    \bottomrule
  \end{tabular}
  \vspace{2pt}
  \footnotesize $n_{\text{broken}}=2323$, $n_{\text{right}}=4118$.
\end{table}

%% file: tables/tab_setB_v3_circular_main_ctrl.tex
\begin{table}[t]
\centering
\caption{\textbf{Circular accuracy under distraction.} Qwen2.5-VL-7B with LoRA arms. Each benchmark column reports mean accuracy over $k\in\{1,2,3,5\}$ distractor images. Pooled columns average across the five benchmarks and report accuracy at $k=0$, at $k=5$, and the mean over $k>0$. $\Delta_{\text{base}}$ is the pooled mean over $k>0$ minus the base model's corresponding mean. Per-$k$ results are shown in Table~\ref{tab:setB_v3_circular}; boldface marks the best model in each column.}
\label{tab:setB_v3_circular_main}
\footnotesize
\setlength{\tabcolsep}{2pt}
\scalebox{0.90}{%
\begin{tabular}{lrrrrrcccc}
\toprule
& \multicolumn{5}{c}{{\bfseries\boldmath Per benchmark} ($k{>}0$ mean)} & \multicolumn{4}{c}{{\bfseries\boldmath Pooled}} \\
\cmidrule(lr){2-6} \cmidrule(lr){7-10}
{\bfseries\boldmath Model} & {\bfseries\boldmath CLEVR} & {\bfseries\boldmath CV-B} & {\bfseries\boldmath MIMIC} & {\bfseries\boldmath WhatsUp} & {\bfseries\boldmath BLINK} & {\bfseries\boldmath $k{=}0$} & {\bfseries\boldmath $k{=}5$} & {\bfseries\boldmath mean} & {\bfseries\boldmath $\Delta_{\text{base}}$} \\
\midrule
Qwen2.5-VL-7B \textit{(base)} & \cellcolor[RGB]{255,244,157}49.9 & \cellcolor[RGB]{238,240,170}65.1 & \cellcolor[RGB]{239,240,169}64.1 & \cellcolor[RGB]{232,238,175}70.6 & \cellcolor[RGB]{255,242,160}47.0 & \cellcolor[RGB]{232,238,175}70.7 & \cellcolor[RGB]{253,244,158}51.4 & \cellcolor[RGB]{244,242,165}59.3 & 0.0 \\
\midrule
\multicolumn{10}{l}{\textit{Ours}} \\
CE only & \cellcolor[RGB]{224,236,181}77.9 & \cellcolor[RGB]{233,239,173}69.1 & \cellcolor[RGB]{247,243,162}56.4 & \cellcolor[RGB]{225,236,180}77.0 & \cellcolor[RGB]{252,244,159}52.6 & \cellcolor[RGB]{230,238,176}72.0 & \cellcolor[RGB]{239,240,169}64.1 & \cellcolor[RGB]{236,240,171}66.6 & +7.3 \\
CE + KL $\lambda{=}0.1$ & \cellcolor[RGB]{226,237,179}76.1 & \cellcolor[RGB]{234,239,173}68.5 & \cellcolor[RGB]{243,241,166}60.6 & \cellcolor[RGB]{224,236,181}77.5 & \cellcolor[RGB]{251,244,159}53.0 & \cellcolor[RGB]{228,237,177}73.7 & \cellcolor[RGB]{239,240,169}63.8 & \cellcolor[RGB]{236,239,172}67.1 & +7.8 \\
CE + KL $\lambda{=}0.5$ & \cellcolor[RGB]{227,237,179}75.0 & \cellcolor[RGB]{233,239,174}69.5 & \cellcolor[RGB]{238,240,170}64.8 & \cellcolor[RGB]{222,236,183}\textbf{79.8} & \cellcolor[RGB]{249,243,161}\textbf{55.0} & \cellcolor[RGB]{228,237,178}74.2 & \cellcolor[RGB]{237,240,171}66.0 & \cellcolor[RGB]{234,239,173}68.8 & +9.5 \\
CE + KL $\lambda{=}1.0$ & \cellcolor[RGB]{219,235,185}\textbf{82.1} & \cellcolor[RGB]{235,239,172}67.8 & \cellcolor[RGB]{237,240,170}\textbf{65.8} & \cellcolor[RGB]{223,236,182}79.0 & \cellcolor[RGB]{251,243,159}53.4 & \cellcolor[RGB]{227,237,178}\textbf{74.8} & \cellcolor[RGB]{236,239,172}\textbf{67.0} & \cellcolor[RGB]{233,239,174}\textbf{69.6} & \textbf{+10.3} \\
CE + KL $\lambda{=}2.0$ & \cellcolor[RGB]{223,236,181}78.4 & \cellcolor[RGB]{234,239,173}68.2 & \cellcolor[RGB]{243,241,165}60.0 & \cellcolor[RGB]{224,236,181}78.1 & \cellcolor[RGB]{250,243,160}54.3 & \cellcolor[RGB]{229,237,177}73.5 & \cellcolor[RGB]{238,240,170}65.2 & \cellcolor[RGB]{235,239,172}67.8 & +8.5 \\
\midrule
\multicolumn{10}{l}{\textit{Reference}} \\
CE + MIMIC mask & \cellcolor[RGB]{222,236,182}79.4 & \cellcolor[RGB]{233,239,174}\textbf{69.8} & \cellcolor[RGB]{246,242,163}57.7 & \cellcolor[RGB]{227,237,178}74.9 & \cellcolor[RGB]{251,244,159}53.3 & \cellcolor[RGB]{229,238,177}72.8 & \cellcolor[RGB]{240,241,168}63.0 & \cellcolor[RGB]{236,239,171}67.0 & +7.7 \\
DTS \textit{(no training)} & \cellcolor[RGB]{244,242,165}59.3 & \cellcolor[RGB]{238,240,170}65.3 & \cellcolor[RGB]{243,241,166}60.3 & \cellcolor[RGB]{230,238,176}72.6 & \cellcolor[RGB]{255,243,159}47.7 & \cellcolor[RGB]{232,238,175}70.7 & \cellcolor[RGB]{248,243,161}55.7 & \cellcolor[RGB]{242,241,166}61.0 & +1.7 \\
\bottomrule
\end{tabular}
}
\vspace{-0.2cm}
\end{table}

%% file: tables/tab_pipeline_steps.tex
\begin{table}[h]
  \centering
  \small
  \setlength{\tabcolsep}{4pt}
  \caption{\textbf{Pipeline steps by benchmark.} All benchmarks except MIMIC, which reuses its native distractors, share the core structure: pool construction, exclusion, vetting, and placement. Key differences are the vetting method and any special handling required.}
  \label{tab:pipeline_steps}
  \begin{tabular}{@{}lp{10cm}@{}}
    \toprule
    \textbf{Benchmark} & \textbf{Pipeline Steps} \\
    \midrule
    CLEVR & \makecell[l]{1. Load questions filtered by category (size/color/count/material) \\
            2. Build attribute index from scene graphs \\
            3. Per-question vetting: exclude renders containing queried attribute \\
            4. Sample $k$ distractors, shuffle with relevant image} \\
    \midrule
    What'sUp & \makecell[l]{1. Parse caption to extract subject + spatial relation \\
               2. Reframe as MCQ: ``Where is/are the X?'' \\
               3. Domain-stratified pool selection (tabletop vs.\ natural) \\
               4. Structural exclusion (own image, other subsets' images) \\
               5. VLM answerability gate (per-candidate check) \\
               6. Sample $k$ distractors, shuffle with relevant image} \\
    \midrule
    BLINK & \makecell[l]{1. Load COCO test2017 as candidate pool for distractors \\
            2. Remove duplicate images \\
            3. VLM answerability gate  \\
            4. Sample $k$ distractors, shuffle with relevant image} \\
    \midrule
    CV-Bench & \makecell[l]{1. Load COCO test2017 as candidate pool for distractors \\ 
              2. Route: COCO test2017 pool (if queried classes are COCO categories) or CV-Bench pool (if not) \\
              3. COCO route: VLM answerability gate  \\
              4. Own pool route: VLM answerability gate \\
              5. Draw colored marker boxes on distractors where the question names one (1,619 rows) \\
              6. Sample $k$ distractors, shuffle with relevant image} \\
    \midrule
    MIMIC & \makecell[l]{1. Load the item's native MIMIC distractors \\
            2. Subsample $k$ of them (no external pool) \\
            3. Shuffle with relevant image} \\
    \bottomrule
  \end{tabular}
\end{table}

%% file: tables/tab_pool_sources.tex
\begin{table}[h]
  \centering
  \small
  \setlength{\tabcolsep}{4pt}
  \caption{Distractor pool sources by benchmark. Pool size indicates the number of candidate images available for distractor selection. What'sUp uses domain-stratified pools to avoid cross-domain artifacts.}
  \label{tab:pool_sources}
  \begin{tabular}{@{}llrl@{}}
    \toprule
    \textbf{Benchmark} & \textbf{Pool Source} & \textbf{Pool Size} & \textbf{Domain Type} \\
    \midrule
    CLEVR & CLEVR val renders & 15,000 & In-domain \\
    \makecell[l]{What'sUp \\ (controlled\_a/b)} & \makecell[l]{controlled\_images + \\ controlled\_clevr} & 829 & In-domain \\
    \makecell[l]{What'sUp \\ (coco\_qa)} & COCO val2017 & 5,000 & In-domain \\
    \makecell[l]{What'sUp \\ (vg\_qa)} & Visual Genome & 4,797 & In-domain \\
    BLINK & COCO test2017 & 40,641 & Matched-domain \\
    CV-Bench & \makecell[l]{COCO test2017 \\ CV-Bench box-free} & \makecell[r]{40,641 \\ 940} & Hybrid \\
    MIMIC & Native per-item distractors (no external pool) & -- & -- \\
    \bottomrule
  \end{tabular}
\end{table}

%% file: tables/tab_vlm_gates.tex
\begin{table}[h]
  \centering
  \small
  \setlength{\tabcolsep}{4pt}
  \caption{\textbf{VLM gate configurations.} Each benchmark uses the strongest vetting guarantee its metadata supports. CLEVR uses exact scene-graph vetting, MIMIC reuses its native distractors, and other benchmarks use VLM-based answerability checks with identical prompts and model.}
  \label{tab:vlm_gates}
  \begin{tabular}{@{}llp{7cm}@{}}
    \toprule
    \textbf{Benchmark} & \textbf{Gate Type} & \textbf{Description} \\
    \midrule
    CLEVR & Ground truth & Scene graphs provide exact attribute vetting; no VLM needed \\
    \midrule
    What'sUp & Answerability & Per-candidate check: ``Does this image contain any of the objects the question asks about, or could this image on its own support answering the question?'' \\
    \midrule
    BLINK & Answerability & Same prompt as What'sUp; applied per candidate image \\
    \midrule
    CV-Bench & \makecell[l]{Answerability} & Same prompt as What'sUp; applied per candidate image \\
    \midrule
    MIMIC & Native (inherited) & Uses MIMIC's own per-item distractors; no additional gate \\
    \bottomrule
  \end{tabular}
\end{table}

%% file: tables/tab_transformations.tex
\begin{table}[h]
  \centering
  \small
  \setlength{\tabcolsep}{4pt}
  \caption{\textbf{Special transformations applied during distractor construction.} Most benchmarks only require size normalization; CV-Bench additionally requires marker drawing to match visual cues in annotated questions.}
  \label{tab:transformations}
  \begin{tabular}{@{}llp{7cm}@{}}
    \toprule
    \textbf{Benchmark} & \textbf{Transformation} & \textbf{Details} \\
    \midrule
    CV-Bench & Marker drawing & \makecell[l]{Extract marker colors from question text \\
                                             Draw matching colored bounding boxes on distractors \\
                                             Match marker pixel mass to relevant image} \\
    \midrule
    BLINK & Duplicate filtering & \makecell[l]{Duplicate detection \\
                                               Remove duplicate images} \\
    \midrule
    All & Resize to match & Distractors resized to relevant image dimensions \\
    \midrule
    All & Alternative variants & \makecell[l]{Randomly sampled distractor images (rnd) \\
                                              Black images (black) \\
                                              Used as control conditions} \\
    \bottomrule
  \end{tabular}
\end{table}

%% file: tables/tab_baseline_rnd.tex
\begin{table}[t]
\centering
\small
\caption{\textbf{Baseline accuracy under random-image distraction.} All three metrics are computed on the same runs: vanilla is the rotation-0 pass, circular requires every rotation to be correct, and mean-rotation is the per-item fraction of rotations passed, so circular $\leq$ mean-rotation holds cell by cell. Vanilla is one member of the rotation average rather than a bound on it, so it can sit marginally below mean-rotation. $k$ is the number of distractor images; $k{=}0$ is the clean single-image baseline and is shared by both distractor kinds. Pooled is the mean of the five benchmark rows.}
\label{tab:baseline_v3_rnd}
\resizebox{\textwidth}{!}{%
\begin{tabular}{l|rrrrr|rrrrr|rrrrr}
\toprule
\textbf{Benchmark} & \multicolumn{5}{c|}{\textbf{Qwen2.5-VL}} & \multicolumn{5}{c|}{\textbf{InternVL3}} & \multicolumn{5}{c}{\textbf{LLaVA-OV}} \\
 & $k=0$ & $k=1$ & $k=2$ & $k=3$ & $k=5$ & $k=0$ & $k=1$ & $k=2$ & $k=3$ & $k=5$ & $k=0$ & $k=1$ & $k=2$ & $k=3$ & $k=5$ \\
\midrule
\multicolumn{16}{l}{\textit{Vanilla accuracy}} \\
CLEVR (size) & \cellcolor[RGB]{213,233,190}88.0 & \cellcolor[RGB]{218,235,186}83.3 & \cellcolor[RGB]{222,236,182}79.1 & \cellcolor[RGB]{227,237,179}75.1 & \cellcolor[RGB]{235,239,172}67.6 & \cellcolor[RGB]{227,237,178}74.6 & \cellcolor[RGB]{231,238,176}71.7 & \cellcolor[RGB]{233,239,174}69.4 & \cellcolor[RGB]{238,240,170}65.4 & \cellcolor[RGB]{240,241,168}63.3 & \cellcolor[RGB]{223,236,182}79.0 & \cellcolor[RGB]{227,237,178}74.9 & \cellcolor[RGB]{229,237,177}73.5 & \cellcolor[RGB]{229,238,177}72.9 & \cellcolor[RGB]{232,238,175}70.6 \\
CV-Bench & \cellcolor[RGB]{223,236,182}79.0 & \cellcolor[RGB]{228,237,178}74.4 & \cellcolor[RGB]{230,238,176}72.4 & \cellcolor[RGB]{231,238,175}71.1 & \cellcolor[RGB]{236,239,172}67.3 & \cellcolor[RGB]{218,235,186}83.1 & \cellcolor[RGB]{221,235,183}80.4 & \cellcolor[RGB]{224,236,181}78.0 & \cellcolor[RGB]{225,236,180}76.7 & \cellcolor[RGB]{230,238,176}72.5 & \cellcolor[RGB]{223,236,181}78.3 & \cellcolor[RGB]{229,238,177}73.2 & \cellcolor[RGB]{231,238,176}71.8 & \cellcolor[RGB]{234,239,173}68.2 & \cellcolor[RGB]{239,240,169}64.1 \\
MIMIC & \cellcolor[RGB]{229,237,177}73.6 & \cellcolor[RGB]{230,238,176}72.3 & \cellcolor[RGB]{231,238,176}71.8 & \cellcolor[RGB]{231,238,175}71.1 & \cellcolor[RGB]{234,239,173}68.9 & \cellcolor[RGB]{226,237,180}76.3 & \cellcolor[RGB]{229,237,177}73.5 & \cellcolor[RGB]{232,238,175}70.6 & \cellcolor[RGB]{233,239,173}69.2 & \cellcolor[RGB]{237,240,170}65.7 & \cellcolor[RGB]{231,238,175}70.9 & \cellcolor[RGB]{233,239,174}69.5 & \cellcolor[RGB]{237,240,171}66.3 & \cellcolor[RGB]{236,240,171}66.4 & \cellcolor[RGB]{238,240,170}65.0 \\
BLINK & \cellcolor[RGB]{236,240,171}66.7 & \cellcolor[RGB]{238,240,170}65.0 & \cellcolor[RGB]{239,240,169}63.8 & \cellcolor[RGB]{242,241,167}61.8 & \cellcolor[RGB]{244,242,165}59.7 & \cellcolor[RGB]{237,240,171}66.3 & \cellcolor[RGB]{236,239,171}66.8 & \cellcolor[RGB]{240,240,168}63.4 & \cellcolor[RGB]{241,241,167}62.3 & \cellcolor[RGB]{245,242,164}58.6 & \cellcolor[RGB]{237,240,171}66.0 & \cellcolor[RGB]{242,241,166}61.0 & \cellcolor[RGB]{243,241,166}60.5 & \cellcolor[RGB]{244,242,165}59.4 & \cellcolor[RGB]{246,242,163}57.6 \\
What'sUp & \cellcolor[RGB]{203,230,198}97.2 & \cellcolor[RGB]{210,232,192}90.6 & \cellcolor[RGB]{212,233,191}88.9 & \cellcolor[RGB]{215,234,188}85.9 & \cellcolor[RGB]{228,237,178}74.1 & \cellcolor[RGB]{204,231,197}95.7 & \cellcolor[RGB]{209,232,193}91.0 & \cellcolor[RGB]{217,234,187}84.3 & \cellcolor[RGB]{223,236,182}78.7 & \cellcolor[RGB]{231,238,175}71.3 & \cellcolor[RGB]{207,232,194}93.2 & \cellcolor[RGB]{217,234,186}84.0 & \cellcolor[RGB]{224,236,181}78.2 & \cellcolor[RGB]{225,236,180}77.2 & \cellcolor[RGB]{241,241,167}62.0 \\
\cmidrule(lr){1-16}
\textit{Pooled} & \cellcolor[RGB]{221,235,184}80.9 & \cellcolor[RGB]{225,236,180}77.1 & \cellcolor[RGB]{227,237,179}75.2 & \cellcolor[RGB]{229,238,177}73.0 & \cellcolor[RGB]{235,239,172}67.5 & \cellcolor[RGB]{222,236,182}79.2 & \cellcolor[RGB]{225,236,180}76.7 & \cellcolor[RGB]{229,238,177}73.1 & \cellcolor[RGB]{232,238,174}70.4 & \cellcolor[RGB]{237,240,171}66.3 & \cellcolor[RGB]{224,236,181}77.5 & \cellcolor[RGB]{230,238,176}72.5 & \cellcolor[RGB]{232,238,174}70.0 & \cellcolor[RGB]{234,239,173}68.8 & \cellcolor[RGB]{239,240,169}63.9 \\
\midrule
\multicolumn{16}{l}{\textit{Circular accuracy}} \\
CLEVR (size) & \cellcolor[RGB]{232,238,175}70.7 & \cellcolor[RGB]{243,241,165}60.1 & \cellcolor[RGB]{249,243,161}54.9 & \cellcolor[RGB]{255,243,158}48.3 & \cellcolor[RGB]{255,234,171}36.4 & \cellcolor[RGB]{255,244,157}50.0 & \cellcolor[RGB]{255,239,163}43.4 & \cellcolor[RGB]{255,235,170}37.7 & \cellcolor[RGB]{255,232,173}34.7 & \cellcolor[RGB]{255,227,179}28.7 & \cellcolor[RGB]{235,239,172}67.9 & \cellcolor[RGB]{240,240,168}63.6 & \cellcolor[RGB]{241,241,167}62.1 & \cellcolor[RGB]{242,241,166}60.9 & \cellcolor[RGB]{245,242,164}59.1 \\
CV-Bench & \cellcolor[RGB]{230,238,176}72.0 & \cellcolor[RGB]{234,239,173}68.6 & \cellcolor[RGB]{237,240,171}66.3 & \cellcolor[RGB]{238,240,170}65.0 & \cellcolor[RGB]{243,241,166}60.4 & \cellcolor[RGB]{222,236,182}79.5 & \cellcolor[RGB]{226,237,180}76.3 & \cellcolor[RGB]{230,238,176}72.5 & \cellcolor[RGB]{231,238,175}71.5 & \cellcolor[RGB]{237,240,170}65.9 & \cellcolor[RGB]{229,238,177}72.8 & \cellcolor[RGB]{236,239,171}66.7 & \cellcolor[RGB]{238,240,170}65.0 & \cellcolor[RGB]{243,241,166}60.3 & \cellcolor[RGB]{249,243,161}55.2 \\
MIMIC & \cellcolor[RGB]{235,239,172}68.0 & \cellcolor[RGB]{238,240,170}64.9 & \cellcolor[RGB]{238,240,169}64.7 & \cellcolor[RGB]{239,240,169}64.1 & \cellcolor[RGB]{241,241,168}62.7 & \cellcolor[RGB]{229,237,177}73.5 & \cellcolor[RGB]{233,239,173}69.1 & \cellcolor[RGB]{235,239,172}68.1 & \cellcolor[RGB]{239,240,169}64.5 & \cellcolor[RGB]{243,241,166}60.7 & \cellcolor[RGB]{236,239,171}66.7 & \cellcolor[RGB]{238,240,170}64.9 & \cellcolor[RGB]{241,241,167}62.3 & \cellcolor[RGB]{243,241,165}60.2 & \cellcolor[RGB]{246,242,163}57.5 \\
BLINK & \cellcolor[RGB]{254,244,157}50.6 & \cellcolor[RGB]{255,244,157}49.5 & \cellcolor[RGB]{255,242,159}47.4 & \cellcolor[RGB]{255,242,160}46.3 & \cellcolor[RGB]{255,240,162}44.7 & \cellcolor[RGB]{244,242,165}59.2 & \cellcolor[RGB]{245,242,164}58.7 & \cellcolor[RGB]{248,243,162}56.3 & \cellcolor[RGB]{248,243,162}55.8 & \cellcolor[RGB]{252,244,158}51.9 & \cellcolor[RGB]{252,244,158}52.3 & \cellcolor[RGB]{255,241,161}46.1 & \cellcolor[RGB]{255,240,162}45.0 & \cellcolor[RGB]{255,238,165}42.2 & \cellcolor[RGB]{255,235,169}38.2 \\
What'sUp & \cellcolor[RGB]{208,232,194}92.2 & \cellcolor[RGB]{220,235,184}81.5 & \cellcolor[RGB]{226,237,180}76.2 & \cellcolor[RGB]{230,238,176}72.0 & \cellcolor[RGB]{252,244,159}52.7 & \cellcolor[RGB]{210,232,192}90.5 & \cellcolor[RGB]{224,236,181}78.2 & \cellcolor[RGB]{233,239,174}69.4 & \cellcolor[RGB]{242,241,166}61.2 & \cellcolor[RGB]{255,243,158}48.7 & \cellcolor[RGB]{215,234,188}85.5 & \cellcolor[RGB]{230,238,176}72.0 & \cellcolor[RGB]{238,240,169}64.6 & \cellcolor[RGB]{245,242,164}58.9 & \cellcolor[RGB]{255,236,167}39.9 \\
\cmidrule(lr){1-16}
\textit{Pooled} & \cellcolor[RGB]{232,238,175}70.7 & \cellcolor[RGB]{238,240,170}64.9 & \cellcolor[RGB]{241,241,167}61.9 & \cellcolor[RGB]{244,242,165}59.1 & \cellcolor[RGB]{253,244,158}51.4 & \cellcolor[RGB]{232,238,175}70.5 & \cellcolor[RGB]{238,240,170}65.2 & \cellcolor[RGB]{243,241,166}60.8 & \cellcolor[RGB]{246,242,163}57.5 & \cellcolor[RGB]{253,244,158}51.2 & \cellcolor[RGB]{234,239,173}69.0 & \cellcolor[RGB]{241,241,168}62.7 & \cellcolor[RGB]{244,242,165}59.8 & \cellcolor[RGB]{247,243,162}56.5 & \cellcolor[RGB]{255,244,157}50.0 \\
\midrule
\multicolumn{16}{l}{\textit{Mean-rotation accuracy}} \\
CLEVR (size) & \cellcolor[RGB]{213,233,190}87.7 & \cellcolor[RGB]{218,235,185}82.9 & \cellcolor[RGB]{222,236,182}79.5 & \cellcolor[RGB]{226,237,179}76.1 & \cellcolor[RGB]{235,239,172}67.7 & \cellcolor[RGB]{227,237,178}74.9 & \cellcolor[RGB]{230,238,176}72.2 & \cellcolor[RGB]{234,239,173}68.6 & \cellcolor[RGB]{237,240,171}66.2 & \cellcolor[RGB]{240,241,168}63.3 & \cellcolor[RGB]{223,236,182}79.0 & \cellcolor[RGB]{227,237,179}75.1 & \cellcolor[RGB]{228,237,177}73.7 & \cellcolor[RGB]{229,238,177}72.8 & \cellcolor[RGB]{231,238,176}71.6 \\
CV-Bench & \cellcolor[RGB]{223,236,181}78.3 & \cellcolor[RGB]{228,237,178}74.1 & \cellcolor[RGB]{230,238,176}72.0 & \cellcolor[RGB]{232,238,175}70.9 & \cellcolor[RGB]{236,239,171}66.9 & \cellcolor[RGB]{218,235,185}82.9 & \cellcolor[RGB]{221,235,183}80.2 & \cellcolor[RGB]{224,236,181}77.3 & \cellcolor[RGB]{226,237,179}76.1 & \cellcolor[RGB]{231,238,176}71.8 & \cellcolor[RGB]{224,236,181}77.3 & \cellcolor[RGB]{229,238,177}72.8 & \cellcolor[RGB]{231,238,175}71.1 & \cellcolor[RGB]{235,239,172}67.6 & \cellcolor[RGB]{240,240,168}63.4 \\
MIMIC & \cellcolor[RGB]{227,237,178}74.6 & \cellcolor[RGB]{230,238,176}72.3 & \cellcolor[RGB]{231,238,176}71.7 & \cellcolor[RGB]{232,238,175}70.8 & \cellcolor[RGB]{233,239,174}69.3 & \cellcolor[RGB]{225,237,180}76.4 & \cellcolor[RGB]{229,238,177}73.2 & \cellcolor[RGB]{231,238,175}71.4 & \cellcolor[RGB]{234,239,173}68.9 & \cellcolor[RGB]{237,240,170}65.5 & \cellcolor[RGB]{232,238,175}70.8 & \cellcolor[RGB]{233,239,173}69.2 & \cellcolor[RGB]{236,239,171}67.0 & \cellcolor[RGB]{236,239,171}66.7 & \cellcolor[RGB]{238,240,169}64.6 \\
BLINK & \cellcolor[RGB]{238,240,170}64.8 & \cellcolor[RGB]{239,240,169}64.1 & \cellcolor[RGB]{241,241,167}61.9 & \cellcolor[RGB]{242,241,167}61.5 & \cellcolor[RGB]{245,242,164}58.8 & \cellcolor[RGB]{236,240,171}66.5 & \cellcolor[RGB]{237,240,170}65.6 & \cellcolor[RGB]{240,240,168}63.3 & \cellcolor[RGB]{241,241,167}62.2 & \cellcolor[RGB]{245,242,164}58.5 & \cellcolor[RGB]{240,241,168}63.2 & \cellcolor[RGB]{243,241,166}60.9 & \cellcolor[RGB]{245,242,164}59.0 & \cellcolor[RGB]{245,242,164}59.0 & \cellcolor[RGB]{248,243,162}55.9 \\
What'sUp & \cellcolor[RGB]{203,230,198}97.1 & \cellcolor[RGB]{210,232,192}90.5 & \cellcolor[RGB]{213,233,190}87.9 & \cellcolor[RGB]{216,234,188}85.4 & \cellcolor[RGB]{228,237,178}74.1 & \cellcolor[RGB]{204,231,197}96.2 & \cellcolor[RGB]{211,233,192}89.8 & \cellcolor[RGB]{217,234,186}83.7 & \cellcolor[RGB]{223,236,181}78.2 & \cellcolor[RGB]{233,239,174}69.8 & \cellcolor[RGB]{207,231,195}93.5 & \cellcolor[RGB]{217,234,187}84.5 & \cellcolor[RGB]{223,236,182}79.0 & \cellcolor[RGB]{226,237,179}75.5 & \cellcolor[RGB]{242,241,167}61.8 \\
\cmidrule(lr){1-16}
\textit{Pooled} & \cellcolor[RGB]{221,235,183}80.5 & \cellcolor[RGB]{225,236,180}76.8 & \cellcolor[RGB]{227,237,178}74.6 & \cellcolor[RGB]{229,238,177}73.0 & \cellcolor[RGB]{235,239,172}67.4 & \cellcolor[RGB]{222,236,182}79.4 & \cellcolor[RGB]{226,237,180}76.2 & \cellcolor[RGB]{229,238,177}72.9 & \cellcolor[RGB]{232,238,174}70.3 & \cellcolor[RGB]{237,240,170}65.8 & \cellcolor[RGB]{225,236,180}76.8 & \cellcolor[RGB]{230,238,176}72.5 & \cellcolor[RGB]{233,239,174}69.9 & \cellcolor[RGB]{234,239,173}68.3 & \cellcolor[RGB]{240,240,168}63.5 \\
\bottomrule
\end{tabular}
}%
\end{table}

%% file: tables/tab_prompting_qwen_rnd.tex
\begin{table}[t]
\centering
\small
\caption{\textbf{Qwen2.5-VL under the relevance prompts, random-image distraction.} Ground Truth names the relevant image; Two-Step selects it then answers; Hard Isolation re-asks with only the selected image in context. $k$ is the number of distractor images. Pooled is the mean of the five benchmark rows.}
\label{tab:prompting_v3_qwen_rnd}
\resizebox{0.85\textwidth}{!}{%
\begin{tabular}{l|rrrr|rrrr|rrrr}
\toprule
\textbf{Benchmark} & \multicolumn{4}{c|}{\textbf{Ground Truth}} & \multicolumn{4}{c|}{\textbf{Two-Step}} & \multicolumn{4}{c}{\textbf{Hard Isolation}} \\
 & $k=1$ & $k=2$ & $k=3$ & $k=5$ & $k=1$ & $k=2$ & $k=3$ & $k=5$ & $k=1$ & $k=2$ & $k=3$ & $k=5$ \\
\midrule
\multicolumn{13}{l}{\textit{Vanilla accuracy}} \\
CLEVR (size) & \cellcolor[RGB]{214,234,189}86.4 & \cellcolor[RGB]{216,234,187}84.6 & \cellcolor[RGB]{218,235,186}83.3 & \cellcolor[RGB]{223,236,182}78.6 & \cellcolor[RGB]{214,234,189}86.4 & \cellcolor[RGB]{216,234,187}84.6 & \cellcolor[RGB]{218,235,186}83.2 & \cellcolor[RGB]{223,236,181}78.3 & \cellcolor[RGB]{213,233,189}87.4 & \cellcolor[RGB]{214,233,189}87.0 & \cellcolor[RGB]{213,233,189}87.3 & \cellcolor[RGB]{213,233,189}87.3 \\
CV-Bench & \cellcolor[RGB]{232,238,174}70.2 & \cellcolor[RGB]{234,239,173}68.9 & \cellcolor[RGB]{233,239,174}69.4 & \cellcolor[RGB]{237,240,171}66.1 & \cellcolor[RGB]{232,238,174}70.2 & \cellcolor[RGB]{234,239,173}68.7 & \cellcolor[RGB]{233,239,173}69.2 & \cellcolor[RGB]{238,240,170}65.4 & \cellcolor[RGB]{224,236,181}78.0 & \cellcolor[RGB]{224,236,181}77.4 & \cellcolor[RGB]{225,236,180}77.1 & \cellcolor[RGB]{227,237,179}75.3 \\
MIMIC & \cellcolor[RGB]{234,239,173}68.4 & \cellcolor[RGB]{236,239,171}67.0 & \cellcolor[RGB]{236,239,171}66.8 & \cellcolor[RGB]{237,240,170}65.7 & \cellcolor[RGB]{234,239,173}68.3 & \cellcolor[RGB]{236,239,171}66.9 & \cellcolor[RGB]{236,239,171}66.9 & \cellcolor[RGB]{237,240,170}65.8 & \cellcolor[RGB]{228,237,178}74.5 & \cellcolor[RGB]{228,237,178}74.3 & \cellcolor[RGB]{228,237,177}73.8 & \cellcolor[RGB]{229,237,177}73.4 \\
BLINK & \cellcolor[RGB]{242,241,166}61.2 & \cellcolor[RGB]{244,242,165}59.4 & \cellcolor[RGB]{245,242,164}58.9 & \cellcolor[RGB]{248,243,162}56.3 & \cellcolor[RGB]{242,241,166}61.3 & \cellcolor[RGB]{244,242,165}59.4 & \cellcolor[RGB]{245,242,164}58.3 & \cellcolor[RGB]{248,243,162}56.3 & \cellcolor[RGB]{238,240,170}65.0 & \cellcolor[RGB]{238,240,170}65.4 & \cellcolor[RGB]{240,241,168}63.3 & \cellcolor[RGB]{240,240,168}63.6 \\
What'sUp & \cellcolor[RGB]{208,232,193}91.8 & \cellcolor[RGB]{210,232,192}90.1 & \cellcolor[RGB]{212,233,190}88.5 & \cellcolor[RGB]{217,234,187}84.5 & \cellcolor[RGB]{210,232,192}90.4 & \cellcolor[RGB]{213,233,190}87.8 & \cellcolor[RGB]{215,234,188}86.3 & \cellcolor[RGB]{223,236,182}78.7 & \cellcolor[RGB]{207,232,194}93.2 & \cellcolor[RGB]{208,232,194}92.1 & \cellcolor[RGB]{209,232,193}91.2 & \cellcolor[RGB]{215,234,188}85.9 \\
\cmidrule(lr){1-13}
\textit{Pooled} & \cellcolor[RGB]{226,237,179}75.6 & \cellcolor[RGB]{228,237,178}74.0 & \cellcolor[RGB]{229,237,177}73.4 & \cellcolor[RGB]{232,238,174}70.2 & \cellcolor[RGB]{227,237,179}75.3 & \cellcolor[RGB]{229,237,177}73.5 & \cellcolor[RGB]{229,238,177}72.8 & \cellcolor[RGB]{234,239,173}68.9 & \cellcolor[RGB]{222,236,183}79.6 & \cellcolor[RGB]{222,236,182}79.2 & \cellcolor[RGB]{223,236,182}78.5 & \cellcolor[RGB]{225,236,180}77.1 \\
\midrule
\multicolumn{13}{l}{\textit{Circular accuracy}} \\
CLEVR (size) & \cellcolor[RGB]{234,239,173}68.2 & \cellcolor[RGB]{239,240,169}64.5 & \cellcolor[RGB]{242,241,166}61.0 & \cellcolor[RGB]{251,244,159}53.0 & \cellcolor[RGB]{234,239,173}68.2 & \cellcolor[RGB]{239,240,169}64.5 & \cellcolor[RGB]{242,241,166}60.9 & \cellcolor[RGB]{251,244,159}52.9 & \cellcolor[RGB]{231,238,175}71.2 & \cellcolor[RGB]{231,238,175}71.3 & \cellcolor[RGB]{232,238,174}70.4 & \cellcolor[RGB]{232,238,174}70.3 \\
CV-Bench & \cellcolor[RGB]{255,238,165}41.7 & \cellcolor[RGB]{255,238,166}41.2 & \cellcolor[RGB]{255,237,166}41.1 & \cellcolor[RGB]{255,235,169}37.7 & \cellcolor[RGB]{255,238,165}41.9 & \cellcolor[RGB]{255,237,166}41.2 & \cellcolor[RGB]{255,237,166}41.0 & \cellcolor[RGB]{255,235,169}38.1 & \cellcolor[RGB]{231,238,175}71.3 & \cellcolor[RGB]{231,238,175}71.0 & \cellcolor[RGB]{232,238,174}70.4 & \cellcolor[RGB]{234,239,173}68.9 \\
MIMIC & \cellcolor[RGB]{255,241,161}45.9 & \cellcolor[RGB]{255,242,160}47.0 & \cellcolor[RGB]{255,242,159}47.2 & \cellcolor[RGB]{255,241,161}45.7 & \cellcolor[RGB]{255,241,161}45.9 & \cellcolor[RGB]{255,242,160}47.0 & \cellcolor[RGB]{255,242,159}47.2 & \cellcolor[RGB]{255,241,161}45.6 & \cellcolor[RGB]{235,239,172}67.9 & \cellcolor[RGB]{235,239,172}67.3 & \cellcolor[RGB]{236,239,172}67.1 & \cellcolor[RGB]{236,239,172}67.2 \\
BLINK & \cellcolor[RGB]{255,225,182}26.2 & \cellcolor[RGB]{255,224,184}24.3 & \cellcolor[RGB]{255,223,184}23.6 & \cellcolor[RGB]{255,222,187}21.4 & \cellcolor[RGB]{255,225,182}25.6 & \cellcolor[RGB]{255,223,184}23.6 & \cellcolor[RGB]{255,223,185}23.3 & \cellcolor[RGB]{255,221,187}20.9 & \cellcolor[RGB]{255,243,158}48.7 & \cellcolor[RGB]{255,244,158}48.9 & \cellcolor[RGB]{255,242,159}47.4 & \cellcolor[RGB]{255,242,160}47.1 \\
What'sUp & \cellcolor[RGB]{221,235,183}80.6 & \cellcolor[RGB]{226,237,179}75.7 & \cellcolor[RGB]{226,237,180}76.2 & \cellcolor[RGB]{232,238,174}70.1 & \cellcolor[RGB]{223,236,182}78.4 & \cellcolor[RGB]{231,238,175}71.2 & \cellcolor[RGB]{230,238,176}72.2 & \cellcolor[RGB]{238,240,169}64.6 & \cellcolor[RGB]{213,233,189}87.3 & \cellcolor[RGB]{216,234,187}84.6 & \cellcolor[RGB]{217,234,187}84.5 & \cellcolor[RGB]{220,235,184}81.0 \\
\cmidrule(lr){1-13}
\textit{Pooled} & \cellcolor[RGB]{252,244,159}52.5 & \cellcolor[RGB]{254,244,157}50.5 & \cellcolor[RGB]{255,244,157}49.8 & \cellcolor[RGB]{255,241,161}45.6 & \cellcolor[RGB]{252,244,158}52.0 & \cellcolor[RGB]{255,244,157}49.5 & \cellcolor[RGB]{255,244,158}48.9 & \cellcolor[RGB]{255,240,162}44.4 & \cellcolor[RGB]{233,239,173}69.3 & \cellcolor[RGB]{234,239,173}68.6 & \cellcolor[RGB]{235,239,172}68.0 & \cellcolor[RGB]{236,239,171}66.9 \\
\midrule
\multicolumn{13}{l}{\textit{Mean-rotation accuracy}} \\
CLEVR (size) & \cellcolor[RGB]{214,233,189}87.0 & \cellcolor[RGB]{216,234,187}84.7 & \cellcolor[RGB]{218,235,186}83.0 & \cellcolor[RGB]{222,236,182}79.3 & \cellcolor[RGB]{214,233,189}86.9 & \cellcolor[RGB]{216,234,187}84.6 & \cellcolor[RGB]{218,235,186}83.0 & \cellcolor[RGB]{222,236,182}79.2 & \cellcolor[RGB]{213,233,190}87.9 & \cellcolor[RGB]{213,233,190}87.8 & \cellcolor[RGB]{213,233,190}87.6 & \cellcolor[RGB]{214,233,189}87.1 \\
CV-Bench & \cellcolor[RGB]{236,240,171}66.4 & \cellcolor[RGB]{237,240,171}66.1 & \cellcolor[RGB]{237,240,170}65.9 & \cellcolor[RGB]{240,240,168}63.4 & \cellcolor[RGB]{236,240,171}66.4 & \cellcolor[RGB]{237,240,170}65.8 & \cellcolor[RGB]{237,240,170}65.5 & \cellcolor[RGB]{240,241,168}63.2 & \cellcolor[RGB]{224,236,181}77.4 & \cellcolor[RGB]{225,236,180}76.9 & \cellcolor[RGB]{225,237,180}76.5 & \cellcolor[RGB]{227,237,178}74.7 \\
MIMIC & \cellcolor[RGB]{235,239,172}68.1 & \cellcolor[RGB]{236,239,172}67.2 & \cellcolor[RGB]{236,240,171}66.6 & \cellcolor[RGB]{237,240,170}65.8 & \cellcolor[RGB]{235,239,172}68.1 & \cellcolor[RGB]{236,239,172}67.0 & \cellcolor[RGB]{236,239,171}66.7 & \cellcolor[RGB]{237,240,170}65.9 & \cellcolor[RGB]{228,237,178}74.2 & \cellcolor[RGB]{228,237,178}74.2 & \cellcolor[RGB]{228,237,177}73.7 & \cellcolor[RGB]{228,237,177}73.9 \\
BLINK & \cellcolor[RGB]{245,242,164}58.2 & \cellcolor[RGB]{248,243,162}56.1 & \cellcolor[RGB]{248,243,162}56.2 & \cellcolor[RGB]{250,243,160}54.5 & \cellcolor[RGB]{246,242,164}58.1 & \cellcolor[RGB]{247,243,162}56.4 & \cellcolor[RGB]{248,243,161}55.6 & \cellcolor[RGB]{250,243,160}54.5 & \cellcolor[RGB]{240,241,168}63.1 & \cellcolor[RGB]{240,241,168}63.0 & \cellcolor[RGB]{241,241,167}62.1 & \cellcolor[RGB]{242,241,166}61.2 \\
What'sUp & \cellcolor[RGB]{209,232,193}91.4 & \cellcolor[RGB]{211,233,191}89.1 & \cellcolor[RGB]{213,233,190}87.8 & \cellcolor[RGB]{215,234,188}85.5 & \cellcolor[RGB]{211,233,192}89.8 & \cellcolor[RGB]{215,234,188}86.2 & \cellcolor[RGB]{216,234,188}85.3 & \cellcolor[RGB]{220,235,184}81.1 & \cellcolor[RGB]{207,232,194}93.1 & \cellcolor[RGB]{209,232,193}91.7 & \cellcolor[RGB]{210,232,192}90.8 & \cellcolor[RGB]{214,233,189}86.7 \\
\cmidrule(lr){1-13}
\textit{Pooled} & \cellcolor[RGB]{228,237,178}74.2 & \cellcolor[RGB]{230,238,176}72.6 & \cellcolor[RGB]{230,238,176}71.9 & \cellcolor[RGB]{233,239,174}69.7 & \cellcolor[RGB]{228,237,177}73.9 & \cellcolor[RGB]{230,238,176}72.0 & \cellcolor[RGB]{231,238,175}71.2 & \cellcolor[RGB]{234,239,173}68.8 & \cellcolor[RGB]{222,236,182}79.1 & \cellcolor[RGB]{223,236,182}78.7 & \cellcolor[RGB]{224,236,181}78.1 & \cellcolor[RGB]{225,236,180}76.7 \\
\bottomrule
\end{tabular}
}%
\end{table}

%% file: tables/tab_content_fraction.tex
\begin{table*}[t]
\centering
\small
\caption{\textbf{Content vs.\ sequence-length contribution to distractor penalty.} $\Delta_{\text{rnd}}$ and $\Delta_{\text{black}}$ are the drops in mean-rotation accuracy from $k{=}0$ to $k{=}5$ distractors. The content fraction measures how much of the penalty is attributable to distractor content rather than sequence length: $(\Delta_{\text{rnd}} - \Delta_{\text{black}}) / \Delta_{\text{rnd}}$.}
\label{tab:content_fraction_v3}
\begin{tabular}{l|rrr|rrr|rrr}
\toprule
& \multicolumn{3}{c|}{\textbf{Qwen2.5-VL}} & \multicolumn{3}{c|}{\textbf{InternVL3}} & \multicolumn{3}{c}{\textbf{LLaVA-OV}} \\
\textbf{Benchmark} & $\Delta_{\text{rnd}}$ & $\Delta_{\text{black}}$ & Content & $\Delta_{\text{rnd}}$ & $\Delta_{\text{black}}$ & Content & $\Delta_{\text{rnd}}$ & $\Delta_{\text{black}}$ & Content \\
\midrule
CLEVR (size) & 20.0 & 1.7 & 91\% & 11.6 & 4.3 & 63\% & 7.3 & $-$1.4 & 119\% \\
CV-Bench & 11.5 & 2.9 & 75\% & 11.1 & 2.2 & 81\% & 14.0 & 3.0 & 79\% \\
MIMIC & 5.2 & 4.0 & 24\% & 10.9 & 6.1 & 44\% & 6.2 & 2.2 & 63\% \\
BLINK & 6.1 & 1.1 & 82\% & 7.9 & 1.5 & 81\% & 7.3 & 2.0 & 72\% \\
What'sUp & 23.0 & 2.6 & 89\% & 26.3 & 6.8 & 74\% & 31.7 & 8.8 & 72\% \\
\midrule
\textbf{Average} & 13.1 & 2.4 & 72\% & 13.6 & 4.2 & 69\% & 13.3 & 2.9 & 81\% \\
\bottomrule
\end{tabular}
\end{table*}

%% file: tables/tab_knockout_spans.tex
\begin{table}[h]
  \centering
  \small
  \setlength{\tabcolsep}{4pt}
  \caption{\textbf{Knockout conditions: source and target spans.} Entries of the additive attention mask at (query $=$ target, key $=$ source) are set to $-\infty$ during prefill, within the designated layer band only. Because the text block follows every image,
  $\mathbf{R} \subset \mathbf{D} \subset \mathbf{P}$.}
  \label{tab:knockout_spans}
  \begin{tabular}{@{}llp{3.1cm}p{5.4cm}@{}}
    \toprule
    \textbf{Cond.} & \textbf{Source} & \textbf{Target} & \textbf{Definition of the target span} \\
    \midrule
    \textbf{P}  & $\textsc{img}_D$ & everything           & every position that is not itself a distractor token \\
    \textbf{MM} &  all preceding  & image  & visual tokens of each image, including the relevant image (MIMIC) \\
    \textbf{D}  & $\textsc{img}_D$ & all text             & every position after the final image: question stem, choices, answer-format instruction, generation scaffold, readout \\
    \textbf{R}  & $\textsc{img}_D$ & readout              & the last prompt position, $L-1$ \\
    \bottomrule
  \end{tabular}
\end{table}

%% file: tables/tab_knockout_ci_middle.tex
\begin{table}[t]
  \centering
  \small
  \setlength{\tabcolsep}{4pt}
  \caption{Attention knockouts with $95\%$ bootstrap intervals (10,000 resamples over items, band \texttt{middle}). \textsc{recov./broken} is the fraction of broken items the knockout fixes. \textsc{broken/right} is the fraction of the already-correct population it damages.}
  \label{tab:knockout-ci}
  \begin{tabular}{llrrlcc}
    \toprule
    \textsc{benchmark} & \textsc{cond.} & $n_{\text{broken}}$ & $n_{\text{right}}$ & \textsc{route} & \textsc{recov./broken} & \textsc{broken/right} \\
    \midrule
    CLEVR & P & 1331 & 1181 & IMG$_D$ $\rightarrow$ everything & 0.54 \tiny{[0.51, 0.56]} & 0.07 \tiny{[0.06, 0.09]} \\
     & MM & 1331 & 1181 & all preceding $\rightarrow$ IMG (MIMIC) & 0.04 \tiny{[0.03, 0.05]} & 0.86 \tiny{[0.83, 0.87]} \\
     & D & 1331 & 1181 & IMG$_D$ $\rightarrow$ all text & 0.44 \tiny{[0.41, 0.47]} & 0.08 \tiny{[0.07, 0.10]} \\
     & R & 1331 & 1181 & IMG$_D$ $\rightarrow$ readout & 0.02 \tiny{[0.02, 0.03]} & 0.04 \tiny{[0.03, 0.05]} \\
    \midrule
    CV-Bench & P & 410 & 1430 & IMG$_D$ $\rightarrow$ everything & 0.58 \tiny{[0.53, 0.63]} & 0.03 \tiny{[0.02, 0.04]} \\
     & MM & 410 & 1430 & all preceding $\rightarrow$ IMG (MIMIC) & 0.19 \tiny{[0.15, 0.22]} & 0.34 \tiny{[0.31, 0.36]} \\
     & D & 410 & 1430 & IMG$_D$ $\rightarrow$ all text & 0.53 \tiny{[0.48, 0.58]} & 0.03 \tiny{[0.02, 0.04]} \\
     & R & 410 & 1430 & IMG$_D$ $\rightarrow$ readout & 0.06 \tiny{[0.04, 0.08]} & 0.01 \tiny{[0.00, 0.02]} \\
    \midrule
    BLINK & P & 89 & 224 & IMG$_D$ $\rightarrow$ everything & 0.55 \tiny{[0.45, 0.65]} & 0.04 \tiny{[0.02, 0.07]} \\
     & MM & 89 & 224 & all preceding $\rightarrow$ IMG (MIMIC) & 0.53 \tiny{[0.43, 0.63]} & 0.46 \tiny{[0.39, 0.52]} \\
     & D & 89 & 224 & IMG$_D$ $\rightarrow$ all text & 0.49 \tiny{[0.39, 0.60]} & 0.05 \tiny{[0.02, 0.08]} \\
     & R & 89 & 224 & IMG$_D$ $\rightarrow$ readout & 0.07 \tiny{[0.02, 0.12]} & 0.02 \tiny{[0.00, 0.04]} \\
    \midrule
    What'sUp (A+B) & P & 339 & 417 & IMG$_D$ $\rightarrow$ everything & 0.66 \tiny{[0.61, 0.71]} & 0.04 \tiny{[0.02, 0.06]} \\
     & MM & 339 & 417 & all preceding $\rightarrow$ IMG (MIMIC) & 0.14 \tiny{[0.10, 0.18]} & 0.50 \tiny{[0.45, 0.55]} \\
     & D & 339 & 417 & IMG$_D$ $\rightarrow$ all text & 0.54 \tiny{[0.48, 0.59]} & 0.04 \tiny{[0.02, 0.06]} \\
     & R & 339 & 417 & IMG$_D$ $\rightarrow$ readout & 0.03 \tiny{[0.01, 0.04]} & 0.01 \tiny{[0.00, 0.02]} \\
    \midrule
    MIMIC & P & 154 & 866 & IMG$_D$ $\rightarrow$ everything & 0.46 \tiny{[0.38, 0.54]} & 0.04 \tiny{[0.03, 0.06]} \\
     & MM & 154 & 866 & all preceding $\rightarrow$ IMG (MIMIC) & 0.10 \tiny{[0.05, 0.15]} & 0.62 \tiny{[0.59, 0.65]} \\
     & D & 154 & 866 & IMG$_D$ $\rightarrow$ all text & 0.45 \tiny{[0.37, 0.53]} & 0.05 \tiny{[0.03, 0.06]} \\
     & R & 154 & 866 & IMG$_D$ $\rightarrow$ readout & 0.05 \tiny{[0.01, 0.08]} & 0.00 \tiny{[0.00, 0.00]} \\
    \bottomrule
  \end{tabular}
\end{table}

%% file: tables/tab_knockout_overlap_middle.tex
\begin{table}[t]
\centering
\caption{\textbf{Item-level overlap between knockout P and knockout D in the middle layers ($9$--$17$).} Percentages are of the broken population. \textbf{P} blocks distractor attention to every token group; \textbf{D} blocks it to text only. The final column is the fraction of D's repairs that P also repairs. Computed from the per-item caches of the same sweep as Table~\ref{tab:knockout-pooled}, so the \textbf{P} and \textbf{D} columns reproduce its recovery rates.}
\label{tab:knockout-overlap}
\small
\begin{tabular}{lrrrrrr}
\toprule
Benchmark & $n$ & \textbf{P} & \textbf{D} & both & D-only & P$\cap$D / D \\
\midrule
CLEVR & 1331 & 54\% & 44\% & 39\% & 4\% & 90\% \\
CV-Bench & 410 & 58\% & 53\% & 49\% & 4\% & 93\% \\
BLINK & 89 & 55\% & 49\% & 47\% & 2\% & 95\% \\
What'sUp (A+B) & 339 & 66\% & 54\% & 52\% & 1\% & 97\% \\
MIMIC & 154 & 46\% & 45\% & 38\% & 6\% & 86\% \\
\midrule
Pooled & 2323 & 56\% & 47\% & 43\% & 4\% & 92\% \\
\bottomrule
\end{tabular}
\end{table}

%% file: tables/tab_knockout_bands.tex
\begin{table}[t]
  \centering
  \small
  \setlength{\tabcolsep}{4pt}
  \caption{\textbf{Knockout recovery by layer band.} Qwen2.5-VL at $k{=}5$ distractors, pooled over five benchmarks at the item level. \textsc{recovery} (fraction of broken items repaired). Bands partition the $28$ decoder layers; \textsc{all} masks every layer. The \textsc{middle} column reproduces Table~\ref{tab:knockout-pooled}. Intervals are 95\% bootstrap CIs over items (10{,}000 resamples).}
  \label{tab:knockout-bands}
  \begin{tabular}{llcccc}
    \toprule
    \textsc{cond.} & \textsc{route} & \textsc{early} & \textsc{middle} & \textsc{late} & \textsc{all} \\
     & & \tiny{0--8} & \tiny{9--17} & \tiny{18--27} & \tiny{0--27} \\
    \midrule
    P & IMG$_D$ $\rightarrow$ everything & 0.34 \tiny{[0.32, 0.36]} & 0.56 \tiny{[0.54, 0.58]} & 0.07 \tiny{[0.06, 0.08]} & 0.76 \tiny{[0.74, 0.77]} \\
    MM & all preceding $\rightarrow$  IMG (MIMIC) & 0.29 \tiny{[0.27, 0.31]} & 0.10 \tiny{[0.09, 0.12]} & 0.03 \tiny{[0.02, 0.04]} & 0.12 \tiny{[0.11, 0.13]} \\
    D & IMG$_D$ $\rightarrow$ all text & 0.19 \tiny{[0.18, 0.21]} & 0.47 \tiny{[0.45, 0.49]} & 0.07 \tiny{[0.06, 0.08]} & 0.57 \tiny{[0.55, 0.59]} \\
    R & IMG$_D$ $\rightarrow$ readout & 0.03 \tiny{[0.03, 0.04]} & 0.03 \tiny{[0.03, 0.04]} & 0.04 \tiny{[0.03, 0.04]} & 0.03 \tiny{[0.03, 0.04]} \\
    \bottomrule
  \end{tabular}
\end{table}

%% file: tables/tab_setB_v3_circular_extra.tex
\begin{table}[t]
\centering
\caption{\textbf{Circular accuracy under distraction, per distractor level.} Qwen2.5-VL-7B with LoRA arms. Circular accuracy: correct under \emph{every} cyclic rotation of the answer options. $k$ is the number of distractor images. One block per benchmark, plus a pooled block averaging the five. The last two columns report the mean over $k>0$ and $\Delta_{\text{base}}$, that mean minus the base model's within the same block. Boldface marks the best value in each column. Summarised in Table~\ref{tab:setB_v3_circular_main}.}
\vspace{3pt}
\label{tab:setB_v3_circular}
\small
\renewcommand{\arraystretch}{0.9}
\setlength{\tabcolsep}{4pt}
\begin{tabular}{lrrrrrcc}
\toprule
& \multicolumn{5}{c}{Distractor images $k$} & \multicolumn{2}{c}{Summary} \\
\cmidrule(lr){2-6} \cmidrule(lr){7-8}
Model & 0 & 1 & 2 & 3 & 5 & Mean $k{>}0$ & $\Delta_{\text{base}}$ \\
\midrule
\multicolumn{8}{l}{\textit{CLEVR-size}} \\
Qwen2.5-VL-7B \textit{(base)} & 70.7 & 60.1 & 54.9 & 48.3 & 36.4 & 49.9 & 0.0 \\
CE only & 82.5 & 79.9 & 78.5 & 77.9 & 75.4 & 77.9 & +28.0 \\
CE + KL $\lambda{=}0.1$ & 82.8 & 79.4 & 77.8 & 75.4 & 71.9 & 76.1 & +26.2 \\
CE + KL $\lambda{=}0.5$ & 79.8 & 77.9 & 75.9 & 74.2 & 72.0 & 75.0 & +25.1 \\
CE + KL $\lambda{=}1.0$ & \textbf{86.7} & \textbf{84.7} & \textbf{82.8} & \textbf{81.4} & \textbf{79.5} & \textbf{82.1} & +32.2 \\
CE + KL $\lambda{=}2.0$ & 83.4 & 80.3 & 80.0 & 77.5 & 75.7 & 78.4 & +28.4 \\
CE + MIMIC mask & 83.6 & 81.4 & 79.9 & 80.0 & 76.1 & 79.4 & +29.4 \\
DTS \textit{(no training)} & 70.7 & 67.1 & 61.4 & 58.0 & 50.6 & 59.3 & +9.3 \\
\midrule
\multicolumn{8}{l}{\textit{CV-Bench}} \\
Qwen2.5-VL-7B \textit{(base)} & 72.0 & 68.6 & 66.3 & 65.0 & 60.4 & 65.1 & 0.0 \\
CE only & 74.7 & 71.0 & 70.0 & 68.6 & 66.8 & 69.1 & +4.0 \\
CE + KL $\lambda{=}0.1$ & 73.9 & 70.9 & 69.7 & 68.0 & 65.6 & 68.5 & +3.5 \\
CE + KL $\lambda{=}0.5$ & 74.3 & 72.2 & 70.0 & 68.8 & \textbf{67.0} & 69.5 & +4.4 \\
CE + KL $\lambda{=}1.0$ & 74.0 & 70.2 & 68.8 & 67.2 & 64.9 & 67.8 & +2.7 \\
CE + KL $\lambda{=}2.0$ & 73.7 & 70.4 & 68.7 & 67.4 & 66.4 & 68.2 & +3.2 \\
CE + MIMIC mask & \textbf{75.4} & \textbf{72.8} & \textbf{70.4} & \textbf{69.4} & 66.7 & \textbf{69.8} & +4.8 \\
DTS \textit{(no training)} & 72.1 & 68.1 & 65.9 & 65.2 & 62.1 & 65.3 & +0.3 \\
\midrule
\multicolumn{8}{l}{\textit{MIMIC}} \\
Qwen2.5-VL-7B \textit{(base)} & 68.0 & 64.9 & 64.7 & 64.1 & 62.7 & 64.1 & 0.0 \\
CE only & 61.2 & 58.7 & 55.9 & 55.7 & 55.3 & 56.4 & -7.7 \\
CE + KL $\lambda{=}0.1$ & 66.5 & 62.9 & 61.3 & 60.9 & 57.5 & 60.6 & -3.5 \\
CE + KL $\lambda{=}0.5$ & \textbf{68.3} & 65.3 & 65.1 & 64.6 & 64.4 & 64.8 & +0.7 \\
CE + KL $\lambda{=}1.0$ & 67.9 & \textbf{66.3} & \textbf{66.1} & \textbf{66.0} & \textbf{64.6} & \textbf{65.8} & +1.6 \\
CE + KL $\lambda{=}2.0$ & 64.5 & 62.3 & 61.3 & 59.1 & 57.4 & 60.0 & -4.1 \\
CE + MIMIC mask & 61.4 & 60.3 & 58.3 & 56.7 & 55.4 & 57.7 & -6.4 \\
DTS \textit{(no training)} & 67.3 & 63.2 & 60.9 & 59.7 & 57.5 & 60.3 & -3.8 \\
\midrule
\multicolumn{8}{l}{\textit{WhatsUp}} \\
Qwen2.5-VL-7B \textit{(base)} & \textbf{92.2} & 81.5 & 76.2 & 72.0 & 52.7 & 70.6 & 0.0 \\
CE only & 85.9 & 80.5 & 79.3 & 76.8 & 71.6 & 77.0 & +6.5 \\
CE + KL $\lambda{=}0.1$ & 88.8 & 81.7 & 78.0 & 77.0 & 73.2 & 77.5 & +6.9 \\
CE + KL $\lambda{=}0.5$ & 90.9 & \textbf{84.6} & \textbf{80.9} & \textbf{79.6} & \textbf{74.0} & \textbf{79.8} & +9.2 \\
CE + KL $\lambda{=}1.0$ & 88.4 & 82.6 & 79.9 & 79.4 & 74.0 & 79.0 & +8.4 \\
CE + KL $\lambda{=}2.0$ & 86.1 & 81.3 & 79.4 & 78.8 & 73.0 & 78.1 & +7.6 \\
CE + MIMIC mask & 88.3 & 82.8 & 76.8 & 73.4 & 66.7 & 74.9 & +4.4 \\
DTS \textit{(no training)} & 92.2 & 81.0 & 73.8 & 72.4 & 63.2 & 72.6 & +2.0 \\
\midrule
\multicolumn{8}{l}{\textit{BLINK}} \\
Qwen2.5-VL-7B \textit{(base)} & 50.6 & 49.5 & 47.4 & 46.3 & 44.7 & 47.0 & 0.0 \\
CE only & 55.5 & 54.7 & 52.4 & 51.9 & 51.3 & 52.6 & +5.6 \\
CE + KL $\lambda{=}0.1$ & 56.3 & 53.9 & 54.5 & 52.9 & 50.6 & 53.0 & +6.0 \\
CE + KL $\lambda{=}0.5$ & 57.9 & \textbf{56.8} & \textbf{55.7} & \textbf{55.2} & 52.4 & \textbf{55.0} & +8.1 \\
CE + KL $\lambda{=}1.0$ & 57.1 & 55.0 & 54.0 & 52.3 & 52.3 & 53.4 & +6.4 \\
CE + KL $\lambda{=}2.0$ & \textbf{59.7} & 55.3 & 54.5 & 54.0 & \textbf{53.4} & 54.3 & +7.4 \\
CE + MIMIC mask & 55.3 & 56.8 & 53.9 & 52.6 & 50.0 & 53.3 & +6.4 \\
DTS \textit{(no training)} & 51.0 & 50.2 & 48.2 & 47.6 & 45.0 & 47.7 & +0.8 \\
\midrule
\multicolumn{8}{l}{\textit{Pooled (5 benchmarks)}} \\
Qwen2.5-VL-7B \textit{(base)} & 70.7 & 64.9 & 61.9 & 59.1 & 51.4 & 59.3 & 0.0 \\
CE only & 72.0 & 69.0 & 67.2 & 66.2 & 64.1 & 66.6 & +7.3 \\
CE + KL $\lambda{=}0.1$ & 73.7 & 69.7 & 68.3 & 66.8 & 63.8 & 67.1 & +7.8 \\
CE + KL $\lambda{=}0.5$ & 74.2 & 71.4 & 69.5 & 68.5 & 66.0 & 68.8 & +9.5 \\
CE + KL $\lambda{=}1.0$ & \textbf{74.8} & \textbf{71.7} & \textbf{70.3} & \textbf{69.3} & \textbf{67.0} & \textbf{69.6} & +10.3 \\
CE + KL $\lambda{=}2.0$ & 73.5 & 69.9 & 68.8 & 67.4 & 65.2 & 67.8 & +8.5 \\
CE + MIMIC mask & 72.8 & 70.8 & 67.9 & 66.4 & 63.0 & 67.0 & +7.7 \\
DTS \textit{(no training)} & 70.7 & 65.9 & 62.0 & 60.6 & 55.7 & 61.0 & +1.7 \\
\bottomrule
\end{tabular}
\end{table}

%% file: tables/tab_kl_prompts_v3.tex
\begin{table}[h]
  \centering
  \small
  \setlength{\tabcolsep}{5pt}
  \caption{\textbf{Circular accuracy (\%) of the KL-clean model ($\lambda{=}1.0$) against the untrained base model}, under the regular, ground truth and two-step prompts. Accuracy is macro-averaged over five benchmark groups weighted equally (What'sUp A and B counted as one). Each cell covers $n{=}9{,}052$ items. Random distractors, \texttt{max\_pixels}${}=1{,}003{,}520$. The relevance prompts have no $k{=}0$ point: selecting among zero distractors is not a task.}
  \label{tab:kl_prompts_v3}
  \begin{tabular}{@{}llrrrrr@{}}
    \toprule
    \textbf{Prompt} & \textbf{Model} & $k{=}0$ & $k{=}1$ & $k{=}2$ & $k{=}3$ & $k{=}5$ \\
    \midrule
    \multirow{3}{*}{Regular}
      & base     & $70.7$ & $64.9$ & $61.9$ & $59.1$ & $51.4$ \\
      & KL-clean & $74.8$ & $71.7$ & $70.3$ & $69.3$ & $67.0$ \\
      & $\Delta$ & $+4.1$ & $+6.8$ & $+8.4$ & $+10.1$ & $+15.7$ \\
    \midrule
    \multirow{3}{*}{Ground Truth}
      & base     & --- & $52.5$ & $50.5$ & $49.8$ & $45.6$ \\
      & KL-clean & --- & $68.6$ & $66.2$ & $65.3$ & $62.3$ \\
      & $\Delta$ & --- & $+16.1$ & $+15.7$ & $+15.4$ & $+16.7$ \\
    \midrule
    \multirow{3}{*}{Two-Step}
      & base     & --- & $52.0$ & $49.5$ & $48.9$ & $44.4$ \\
      & KL-clean & --- & $67.9$ & $65.6$ & $64.6$ & $61.2$ \\
      & $\Delta$ & --- & $+16.0$ & $+16.1$ & $+15.7$ & $+16.8$ \\
    \bottomrule
  \end{tabular}
\end{table}